\documentclass[letterpaper]{article} 
\usepackage{aaai2026}  
\usepackage{times}  
\usepackage{helvet}  
\usepackage{courier}  
\usepackage[hyphens]{url}  
\usepackage{graphicx} 
\usepackage{natbib}  
\usepackage{caption} 
\usepackage{algorithm}
\usepackage{algorithmic}

\usepackage{newfloat}
\usepackage{listings}
\DeclareCaptionStyle{ruled}{labelfont=normalfont,labelsep=colon,strut=off} 
\floatstyle{ruled}
\newfloat{listing}{tb}{lst}{}
\floatname{listing}{Listing}
\usepackage{booktabs}
\usepackage{amsmath}
\usepackage{multirow}
\usepackage{amssymb}
\usepackage{pifont}
\usepackage{array} 
\newcolumntype{P}[1]{>{\centering\arraybackslash}p{#1}}

\title{Driving with Regulation: Trustworthy and Interpretable Decision-Making for
Autonomous Driving with Retrieval-Augmented Reasoning}

\author {
    Tianhui Cai\equalcontrib,
    Yifan Liu\equalcontrib,
    Zewei Zhou,
    Haoxuan Ma,
    Seth Z. Zhao,\\
    Zhiwen Wu,
    Xu Han,
    Zhiyu Huang\thanks{Corresponding author. Email: \texttt{zhiyuh@ucla.edu}},
    Jiaqi Ma
}

\affiliations {
    University of California, Los Angeles \\
}

\usepackage{bibentry}

\begin{document}

\maketitle

\begin{abstract}
Understanding and adhering to traffic regulations is essential for autonomous vehicles to ensure safety and trustworthiness. However, traffic regulations are complex, context-dependent, and differ between regions, posing a major challenge to conventional rule-based decision-making approaches. We present an interpretable, regulation-aware decision-making framework, \textbf{DriveReg}, which enables autonomous vehicles to understand and adhere to region-specific traffic laws and safety guidelines. The framework integrates a Retrieval Augmented Generation (RAG)-based Traffic Regulation Retrieval Agent, which retrieves relevant rules from regulatory documents based on the current situation, and a Large Language Model (LLM)-powered Reasoning Agent that evaluates actions for legal compliance and safety. Our design emphasizes interpretability to enhance transparency and trustworthiness. To support systematic evaluation, we introduce \textbf{DriveReg Scenarios Dataset}, a comprehensive dataset of driving scenarios across Boston, Singapore, and Los Angeles, with both hypothesized text-based cases and real-world driving data, specifically constructed and annotated to evaluate models’ capacity for regulation understanding and reasoning. We validate our framework on the DriveReg Scenarios Dataset and real-world deployment, demonstrating strong performance and robustness across diverse environments.
\end{abstract}

\begin{links}
\small
\link{Code}{https://github.com/Vickycth/driving-with-regulation}
\link{Extended version}{https://arxiv.org/abs/2410.04759}
\end{links}

\section{Introduction}
Autonomous driving technologies have advanced significantly in recent years, showing potential to improve safety and efficiency \cite{zhao2023autonomous, han2024foundation, huang2025gen, huang2023gameformer,li2023pretraining,li2021metadrive,li2023scenarionet}.  However, the societal deployment of autonomous vehicles (AVs) depends not only on technical progress in perception or control, but also on their ability to operate legally, transparently, and in accordance with human expectations~\cite{kubica2022AVLL, USDOT2016FAVP}. Understanding and complying with traffic regulations is vital for safety and public trust~\cite{MEHDIPOUR2023110692}. Therefore, beyond technical robustness, AVs must ensure legally compliant and interpretable decision-making to enable trustworthy, socially-aligned deployment.

Despite its importance, understanding traffic rules remains a major challenge for AVs. Regulations are complex, context-dependent, and vary across regions, and small contextual changes can lead to different rule interpretations. While prior work encodes traffic rules via hand-crafted logic~\cite{xiao2021rule, manas2022robust, sun2022lawbreaker}, such approaches lack scalability and adaptability. Moreover, traffic rules span strict legal codes and flexible safety guidelines, demanding diverse reasoning strategies. This semantic richness exceeds the capabilities of conventional models, but recent advances in Large Language Models (LLMs) offer a promising solution through their ability to interpret natural language and perform context-aware reasoning.

In addition to system development challenges, there is no comprehensive benchmark for evaluating traffic rules understanding in AV decision-making. Datasets like Argoverse 2 \cite{wilson2023argoverse} and nuPlan \cite{nuplan} focus on perception or planning, without annotations linking actions to specific rules. Existing traffic rule-aware datasets either lack real-world scenarios \cite{Li2024VioHawk, Deng2025TARGET}, or cover only a limited subset of regulations \cite{sun2022lawbreaker, Deng2025TARGET, Li2024VioHawk}.

To address these limitations, we introduce \textbf{DriveReg}, an interpretable, traffic regulation-aware decision-making framework for AVs, along with the \textbf{DriveReg Scenarios Dataset}, a benchmark of 500 scenarios spanning Boston, Singapore, and Los Angeles. DriveReg integrates a Traffic Regulation Retrieval (TRR) Agent built upon Retrieval-Augmented Generation (RAG), with an LLM-based Reasoning Agent. Given a driving scenario, the system retrieves relevant traffic rules from an extensive collection of regulatory documents applicable to the city and evaluates each candidate action on two levels: (1) \textit{\textbf{Compliance}}, whether it satisfies mandatory legal constraints; and (2) \textit{\textbf{Safety}}, whether it also adheres to non-mandatory guidelines. To ensure transparency, DriveReg outputs reasoning steps with references to the specific rules used in the decision. The \textbf{DriveReg Scenarios Dataset} includes 360 hypothesized cases (120 per city), designed to cover all categories defined in each region’s regulatory documents, and 140 real-world scenarios sourced from the nuScenes dataset and our in-house dataset. Each scenario is annotated with the relevant traffic rules and labeled at the action level for compliance and safety, enabling fine-grained evaluation across different cities.
The main contributions of our paper are summarized as follows:
\begin{itemize}
    \item We propose \textbf{DriveReg}, an interpretable, LLM-driven decision-making framework for autonomous driving that integrates a Traffic Regulation Retrieval Agent and a Reasoning Agent to assess action-level compliance and safety, enabling regulation-adherent and region-aware decision-making.

    \item We introduce the \textbf{DriveReg Scenarios Dataset}, and a benchmark of 500 driving scenarios (hypothesized and real-world) from Boston, Singapore, and Los Angeles, annotated with retrieved traffic rules and compliance/safety labels for action-level evaluation.

    \item We demonstrate the effectiveness, robustness, and regional generalization of DriveReg through extensive experiments on the DriveReg Scenarios Dataset and validate the DriveReg framework in real-world deployment to assess its practical performance.
\end{itemize}

\section{Related Work}

\subsection{Traffic Regulation in Autonomous Driving}

Various techniques have been applied to integrate traffic regulations into autonomous driving systems. Early approaches included rule-based systems \cite{li2023knowledgedrivenautonomousdriving} and finite state machines \cite{Bae2020FiniteSM}, which encoded traffic laws through explicit if-then rules or state transitions. To handle complex scenarios, more sophisticated methods emerged, such as using behavior trees~\cite{9030183}, and formal methods~\cite{9304549}. However, these methods often struggled with the ambiguity and regional variations of real-world traffic rules.

Recently, Large Language Models (LLMs) have demonstrated remarkable capabilities in understanding natural language and interpreting complex scenarios~\cite{wen2024enhancingsociallyawarerobotnavigation, zhang2024large, sima2023drivelm, malla2023drama, wei2024editable}. LLMs can process traffic rules in a more flexible and context-aware manner. For example, LLaDA \cite{li2024drivinglargelanguagemodel} utilizes LLMs to interpret traffic rules from local handbooks, while Agent-Driver \cite{mao2024languageagentautonomousdriving} incorporates traffic rules into an LLM-based cognitive framework. However, ensuring LLMs accurately apply relevant traffic rules without hallucinations remains a key challenge.

\subsection{Retrieval-Augmented Generation}
Retrieval-Augmented Generation (RAG) \cite{lewis2021retrieval} addresses LLM hallucinations by combining neural retrieval with sequence-to-sequence generation based on relevant documents. Recent studies \cite{pmlr-v162-borgeaud22a, nakano2022webgpt} have demonstrated significant improvements in LLM accuracy and factual correctness across domains. For autonomous driving systems, RAG's dynamic retrieval capability \cite{lewis2021retrieval} enables real-time access to region-specific traffic rules, while its enhanced factual grounding \cite{pmlr-v162-borgeaud22a} reduces the risk of fabricating or misapplying regulations. Additionally, RAG's ability to handle complex contextual \cite{nakano2022webgpt} is well-suited for interpreting nuanced traffic regulations with multiple conditions, and its inherent transparency improves decision-making interpretability, crucial for regulatory compliance and public trust.

\subsection{Decision-Making of Autonomous Driving}
Decision-making methods for autonomous driving have evolved from rule-based \cite{wang2021decision} to learning-based methods \cite{kiran2021deep}, which offer greater adaptability in dynamic environments. Typical learning-based approaches include imitation learning~\cite{bansal2018chauffeurnet, tang2023personalized} and reinforcement learning~\cite{yuan2024evolutionary}. More recently, GPT-Driver \cite{mao2023gpt} has reformulated motion planning as a language modeling problem. However, the integration of diverse semantic traffic rules into decision-making using a unified model remains underexplored.

\begin{figure*}[!t]
    \centering
    \medskip
    \includegraphics[width=0.96\linewidth]{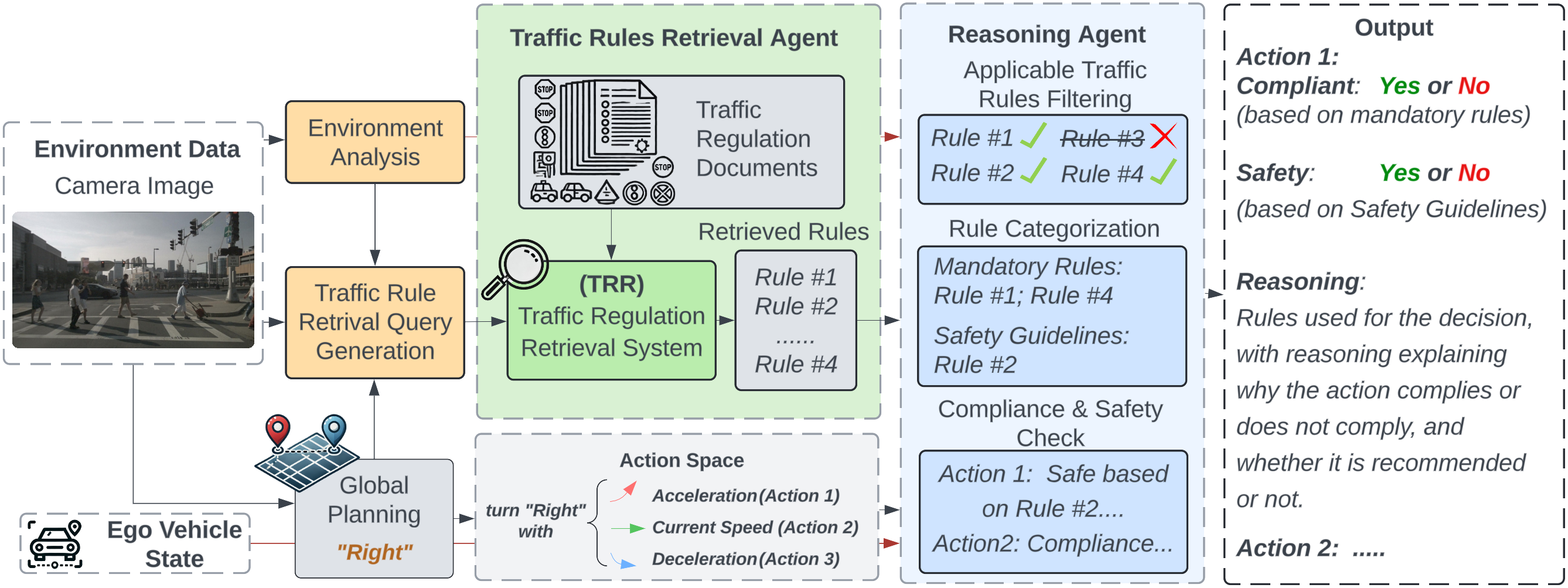}
    \caption{\textbf{Overview of Driving with Regulation (DriveReg) Framework}. The framework consists of two main components: the Traffic Rules Retrieval Agent and the Reasoning Agent. The Traffic Rules Retrieval Agent retrieves relevant rules from traffic regulation documents based on the generated traffic rule retrieval query. The Reasoning Agent then identifies the applicable rules from the retrieved set and performs compliance and safety checks based on those applicable rules.}
    \label{fig:pipeline}
\end{figure*}

\section{Method}

\begin{figure*}[t]
    \centering
    \medskip
    \includegraphics[width=0.96\linewidth]{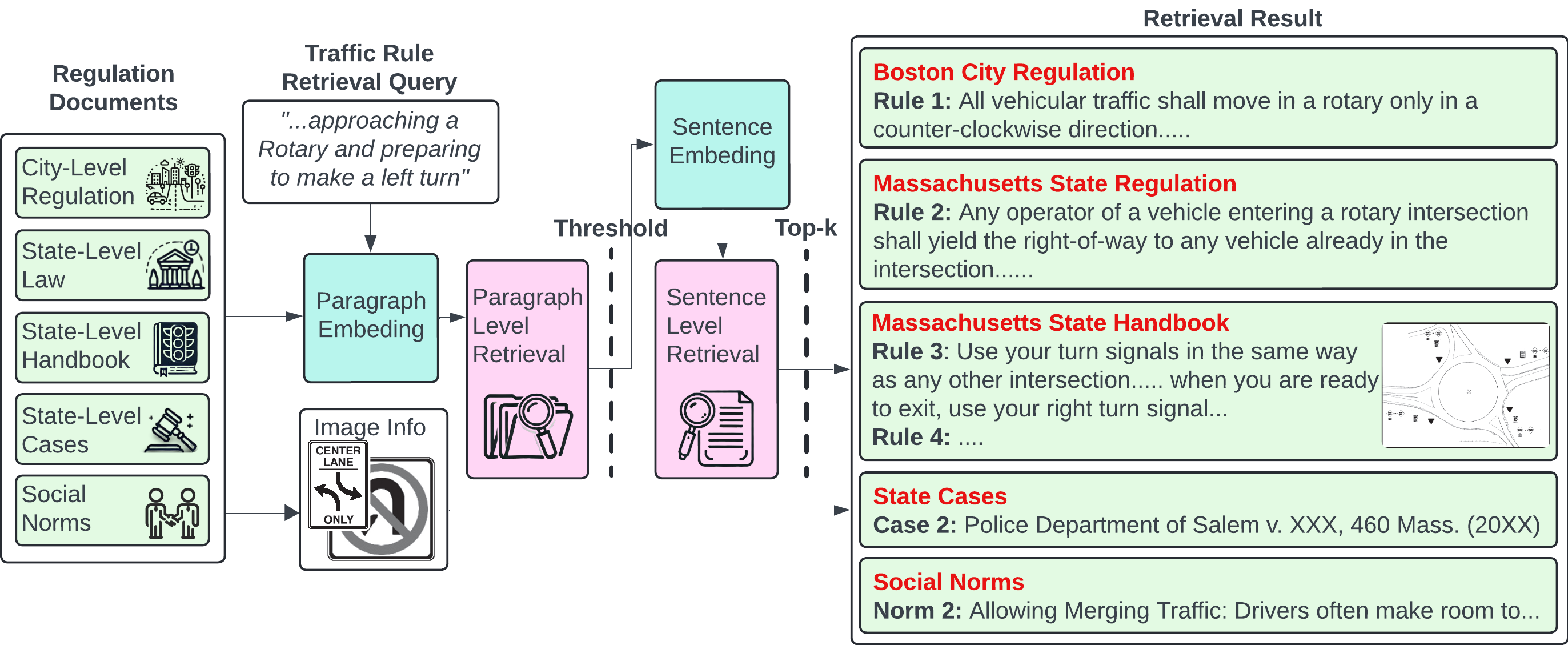}
    \caption{\textbf{Illustration of the proposed Traffic Regulation Retrieval (TRR) Agent.} The retrieval results are obtained through the similarity score between scene description and well-curated regulation documents with a pre-defined relevance metric.}
    \label{fig:trr_architecture}
\end{figure*}

\subsection{Overview}
Our proposed method, as shown in Fig. \ref{fig:pipeline}, comprises two main components: a \textbf{Traffic Rules Retrieval (TRR) Agent} that retrieves relevant traffic rules from regulation documents using a retrieval query, and a \textbf{Reasoning Agent} that assesses the traffic rule adherence of each action in the proposal set based on environment information, ego vehicle's state, and retrieved traffic rules.

To support traffic rule retrieval, we perform an environment analysis using a Vision-Language Model (VLM), denoted as $\mathcal{V}$. Given multi-view images $I$ from the ego vehicle and navigation intent $g \in \{\text{Left, Right, Forward}\}$, $\mathcal{V}$ follows a structured Chain-of-Thought (CoT) reasoning process: it first provides a high-level overview of the driving scene, including road layout and general traffic structure. It then performs a more focused analysis of critical elements, such as traffic signs, road users, and lane markings, particularly those relevant to the intended maneuver $g$. Finally, the VLM generates a concise natural language query that summarizes the current scenario for retrieving relevant traffic rules. We denote this reasoning process as $\mathcal{V}(I, g) \rightarrow (c, q)$, where $c$ is the structured description of the scenario and $q$ is the retrieval query provided to the TRR Agent. An example of this environment analysis output is shown in Fig.~\ref{fig:result_hypo}.
 
For decision-making, we extract an action proposal set containing possible actions from the Action Space $\mathcal{A}$ based on the Global Planning output $g$. For simplicity and to maintain the focus of this work on traffic rule adherence, the Action Space $\mathcal{A}$ only consists of a predefined set of actions: turning right, turning left, going forward (with current speed, acceleration, or deceleration), changing lane to the left, and changing lane to the right. The candidate set $\mathcal{A}_{\text{cand}}$ is obtained by selecting actions from $\mathcal{A}$ that align with the current global intent $g$. For instance, if $g = \text{Left}$, then $\mathcal{A}_{\text{cand}} = $ \{turning left with current speed, turning left with an acceleration, turning left with a deceleration\}. 

Given the retrieval query $q$, the Traffic Rules Retrieval Agent ($\mathrm{TRR}$) fetches a set of relevant traffic rules $\mathcal{R}_q$ from a collection of traffic regulatory documents $\mathcal{D}$. The Traffic Reasoning Agent ($\mathrm{TRA}$) evaluates each candidate action $a \in \mathcal{A}_{\text{cand}}$ for both compliance and safety, leveraging the structured scene description $c$ and the set of retrieved traffic rules $\mathcal{R}_q$, obtained by querying the Traffic Rules Retrieval Agent with $q$. Formally, the agents perform:
\begin{equation}
\mathrm{TRR}(q; \mathcal{D}) \rightarrow \mathcal{R}_q, \, 
\mathrm{TRA}(a, c, \mathcal{R}_q) \rightarrow (l_c, l_s),
\end{equation}
where $l_c, l_s \in \{0, 1\}$ are binary outputs for compliance and safety, respectively, with $1$ indicating compliant or safe, and $0$ otherwise.

The final decision of the system is the set of actions that are both compliant and safe:
\begin{equation}
\mathcal{A}^* = \left\{ a \in \mathcal{A}_{\text{cand}} \ \middle| \ l_c(a) = 1 \ \land\ l_s(a) = 1 \right\}.
\end{equation}

\subsection{Traffic Rules Retrieval Agent}
To enhance the model's understanding of local traffic rules and norms, and to fully consider all the related rules from available sources, we developed the Traffic Regulation Retrieval Agent, which employs a two-level retrieval strategy as illustrated in Fig. \ref{fig:trr_architecture}. 

Since different regions have varying sources of traffic rules, we use the United States as an example to illustrate the comprehensive collection of regulatory documents $\mathcal{D}$ used by the TRR Agent, including state-level traffic laws, official state driving manuals (which also include safety guidelines), city-level regulations, relevant court cases that establish legal precedent, and widely accepted traffic norms.

During the retrieval process, we first encode both the traffic rule retrieval query $q$ and the regulatory documents $\mathcal{D}$ into dense vector representations using a sentence embedding model $\mathcal{E}(\cdot)$. Each paragraph $p_i \in \mathcal{D}$ is embedded as $\mathbf{v}_{p_i} = \mathcal{E}(p_i)$, and the query is embedded as $\mathbf{v}_q = \mathcal{E}(q)$. We utilize FAISS~\cite{johnson2017billion} to efficiently retrieve a top-$k$ candidate set of paragraphs based on vector similarity:
\begin{equation}
\mathcal{P}_q^{\text{cand}} = \texttt{TopK}_{\text{FAISS}}\left( \mathbf{v}_q, \left\{ \mathbf{v}_{p_i} \right\}_{i=1}^{|\mathcal{D}|} \right).
\end{equation}

We retain only paragraphs with similarity scores exceeding a threshold $\tau_p$:
\begin{equation}
\mathcal{P}_q = \{ p_i \in \mathcal{P}_q^{\text{cand}} \mid \text{sim}(\mathbf{v}_q, \mathbf{v}_{p_i}) \geq \tau_p \}.
\end{equation}

To address the sparsity and granularity issues that arise from long-form regulatory text, we further refine the retrieval by segmenting each paragraph in $\mathcal{P}_q$ into individual sentences and performing sentence-level retrieval. Specifically, each paragraph $p \in \mathcal{P}q$ is segmented into a set of sentences ${s_j}$, each encoded as a sentence-level embedding $\mathbf{v}_{s_j} = \mathcal{E}'(s_j)$, where $\mathcal{E}’$ is a lighter-weight sentence encoder. A second FAISS-based similarity search is then conducted to identify the most relevant sentences to $\mathbf{v}'_{q} = \mathcal{E}'(q)$:

\begin{equation}
\mathcal{R}_q = \texttt{TopK}_{\text{FAISS}}( \mathbf{v'}_q, \{ \mathbf{v}_{s_j} \}_{j=1}^{|\mathcal{S}|} ),
\end{equation}
where $\mathcal{S} = \bigcup_{p \in \mathcal{P}_q} \text{Sentences}(p)$.

Our two-level retrieval strategy combines paragraph-level retrieval for broad coverage with sentence-level refinement for precision, yielding concise, context-aligned rule snippets while improving efficiency and semantic relevance. By anchoring the Reasoning Agent in retrieved regulations, the TRR Agent helps mitigate hallucinations of LLMs and enhance decision reliability and transparency.

\subsection{Reasoning Agent}
The Reasoning Agent is responsible for determining whether each action in the proposal set complies with traffic rules, leveraging an LLM (e.g., GPT-4o) with CoT prompting and few-shot learning. The Reasoning Agent receives three key inputs: (1) the current environment information from the environment analysis $c$, (2) the ego vehicle’s action proposal set $\mathcal{A}_{\text{cand}}$, and (3) a set of retrieved traffic rules from TRR Agent $\mathcal{R}_q$.

In the reasoning pipeline, the agent first filters $\mathcal{R}_q$ to identify rules relevant to the current scene and the ego vehicle’s intended maneuver. The filtered rules are categorized into two types: \textbf{mandatory rules}, which must be satisfied to ensure legal compliance, and \textbf{safety guidelines}, which represent best practices not legally binding but recommended for safe operation.

\begin{figure*}[t]
    \centering
    \medskip
    \includegraphics[width=\linewidth]{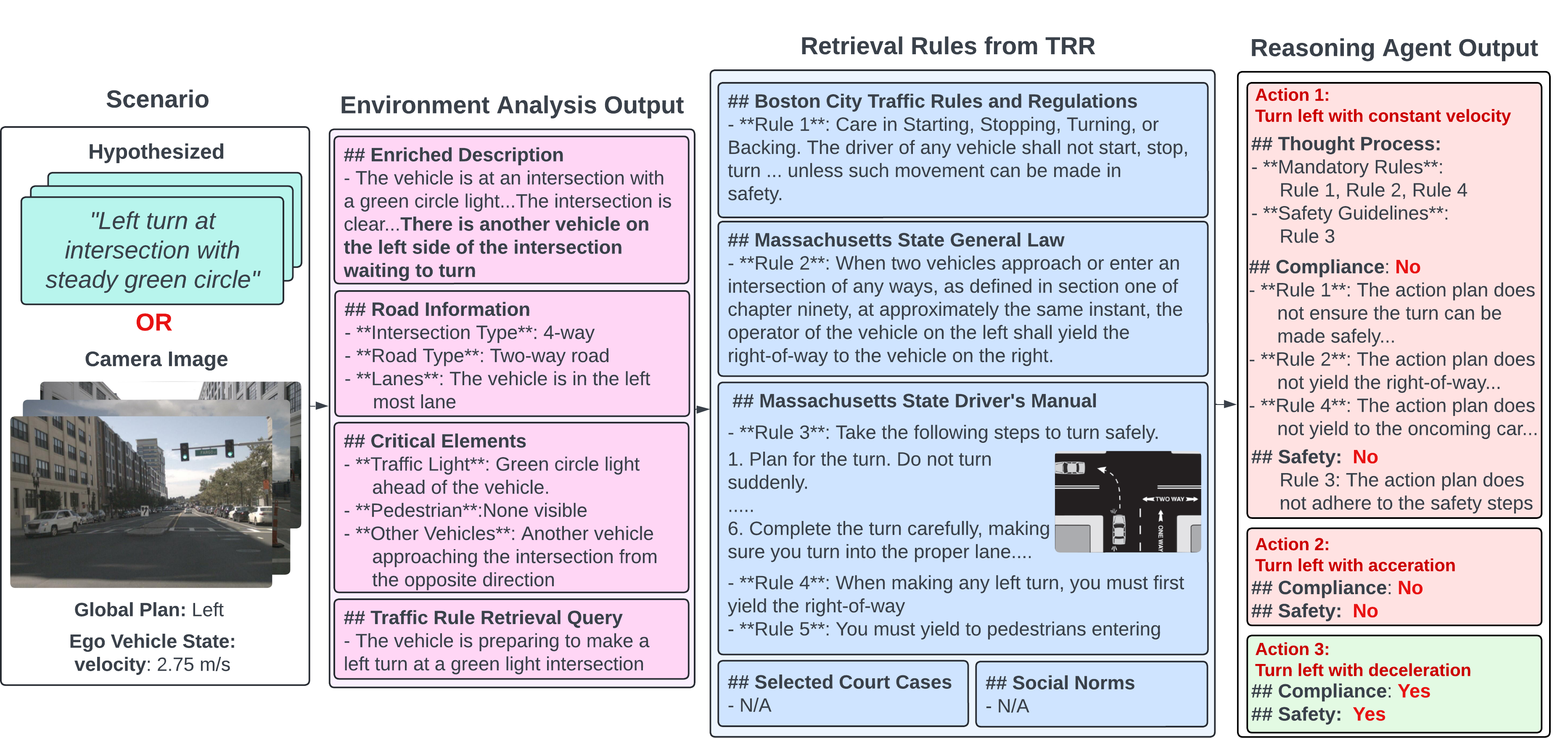}
    \caption{\textbf{Pipeline of processing the selected scenario.} The correct action is labeled in \textbf{green} background.}
    \label{fig:result_hypo}
\end{figure*}

For each action $a \in \mathcal{A}_{\text{cand}}$, the Reasoning Agent first outputs a compliance indicator $l_c = 1$ if the action does not violate any relevant mandatory rule in $\mathcal{R}_q$, and $l_c = 0$ otherwise. It then produces a safety indicator $l_s = 1$ if the action satisfies both the mandatory rules and any retrieved safety guidelines, and $l_s = 0$ if any of these are violated. Alongside each binary indicator, the Reasoning Agent provides a concise explanation referencing the specific rules involved, clarifying why the action is considered compliant or non-compliant, safe or unsafe. This interpretability is critical for transparency in decision-making. The framework then selects the actions that are marked as both compliant and safe as the final output for decision-making. An example output of the Reasoning Agent is shown in Fig. \ref{fig:result_hypo}.

\begin{figure}[h]
    \centering
    \medskip
    \includegraphics[width=\linewidth]{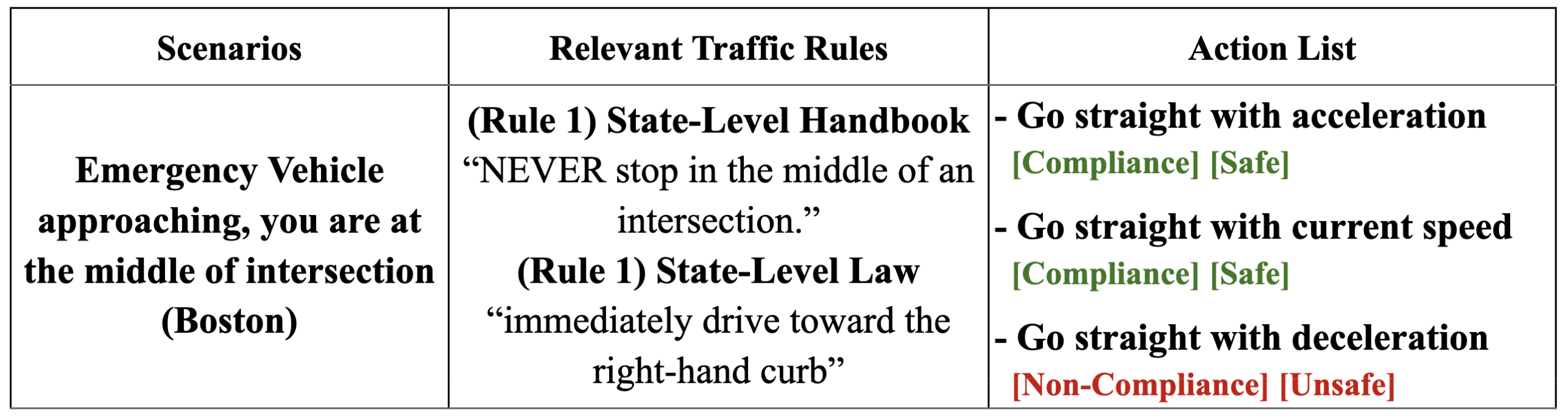}
    \caption{Example from the DriveReg Scenarios Dataset showing scenario descriptions, relevant traffic rules, and action-level compliance/safety annotations.}
    \label{fig:dataset_examples}
\end{figure}

\begin{figure*}[tb]
    \centering
    \medskip
    \includegraphics[width=\linewidth]{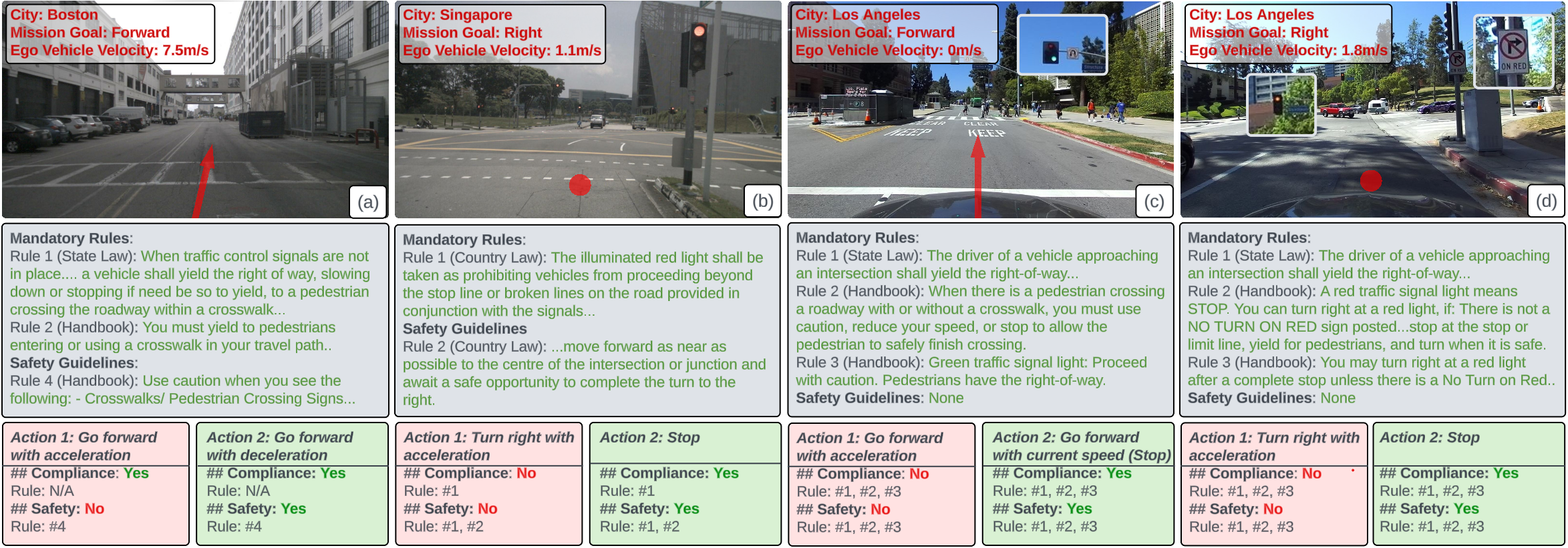}
    \caption{\textbf{Inference results from the DriveReg (Real-World) Scenarios Dataset.} Actions in green boxes represent the final decision-making outputs, which are both compliant and safe. Results demonstrate that our framework successfully retrieves relevant traffic rules and interprets and assesses actions based on those rules. The correct action is labeled in a green background. (Due to space constraints, some actions and reasoning details are omitted.)}
    \label{fig:real_world_boston_singapore_la}
\end{figure*}

\subsection{DriveReg Scenarios Dataset}

\begin{table}[t]
\scriptsize	
\centering
\setlength{\tabcolsep}{1mm}
\begin{tabular}{l|cccc}
\toprule
Dataset & \# Rules & \# Scenarios & Real-World & Multi-Region \\
\midrule
VioHawk~\cite{Li2024VioHawk} & 27 & 42 & $\times$ & $\times$\\
TARGET~\cite{Deng2025TARGET} & 54 & 284 & $\times$ & $\times$ \\
LawBreaker~\cite{sun2022lawbreaker} & 24 & 173 & $\times$ & $\times$  \\
\textbf{DriveReg} & \textbf{562} & \textbf{500} & \checkmark & \checkmark \\
\bottomrule
\end{tabular}
\caption{Comparison of existing traffic rule-aware driving datasets. DriveReg provides the most comprehensive coverage with real-world data, multi-region scenarios, and fine-grained compliance and safety labels.}
\label{tab:dataset_comparison}
\end{table} 

We introduce the \textbf{DriveReg Scenarios Dataset}, the first dataset designed to evaluate models on understanding and reasoning over traffic regulations in diverse driving scenarios. It contains 360 hypothesized text-based scenarios and 140 real-world cases with front-camera images drawn from the nuScenes dataset and data collected during our deployment testing. Hypothesized scenarios offer greater diversity, while real-world data evaluates the framework’s practical performance in real driving conditions. For each city, we preprocess comprehensive traffic regulation sources, and each scenario is annotated with the relevant traffic rules and the two-level decision labels indicating whether an action is (1) legally compliant and (2) safe, based on the applicable regulations. An example is shown in Fig.~\ref{fig:dataset_examples}, with additional cases provided in the extended version, and comparison with existing traffic rule-aware datasets is shown in Table~\ref{tab:dataset_comparison}.

\noindent \textbf{Traffic Regulation Sources.} 
We use extensive traffic regulation sources for the three cities included in the DriveReg Scenarios Dataset, covering city ordinances, state or national laws, official driver manuals, and selected legal cases or norms. For the Boston region, this includes the Boston City Traffic Rules and Regulations, the Massachusetts General Laws, the Massachusetts Driver’s Manual, and twelve selected state court cases on traffic violations. We also include ten representative U.S. driving norms that reflect lawful behavior without explicit violations. More details and source descriptions for Singapore and Los Angeles are provided in the extended version.

\noindent \textbf{Scenario Collection and Annotation Process.}
\noindent \textit{1) Hypothesized Scenarios.} We construct 120 hypothesized scenarios per city, including 100 \textbf{normal} and 20 \textbf{hard} cases. \textbf{Normal scenarios} are designed to cover all regulatory categories and document sections, reflecting common traffic situations such as right-of-way decisions and pedestrian interactions. Each involves 1-2 straightforward rules to evaluate basic regulation understanding. \textbf{Hard scenarios} include cases with region-specific regulations and complex conditions involving three or more applicable rules, such as school-bus handling at intersections with malfunctioning signals.

The distribution of hypothesized scenarios for Boston across regulatory categories is shown in Table~\ref{tab:hypo_scenarios_classification}, demonstrating comprehensive coverage across key regulatory domains. Similar distributions were constructed for Singapore and Los Angeles and are provided in the extended version.

\begin{table}[h]
\centering
\scriptsize 
\setlength{\tabcolsep}{3mm}
\begin{tabular}{lc|lc}
\toprule
\textbf{Category} & \textbf{Count} & \textbf{Category} & \textbf{Count} \\
\midrule
Right-of-Way Rules     & 20 & Special Driving Situations & 11 \\
Intersections          & 15 & Rules for Passing         & 8  \\
Pedestrians            & 15 & Traffic Signs \& Signals  & 8  \\
Pavement Markings      & 10 & Roadway Construction      & 5  \\
Service Vehicles       & 8  &  Hard Cases         & 20 \\
\bottomrule
\end{tabular}
\caption{Distribution of Hypothesized Scenarios in Boston.}
\label{tab:hypo_scenarios_classification}
\end{table}

\noindent \textit{2) Real-World Scenarios.} To evaluate models' performance in real-world driving contexts, we selected and annotated 140 scenarios based on two sources: (1) nuScenes dataset \cite{caesar2020nuscenes}, which captures urban driving scenes in Boston and Singapore, and (2) an in-house dataset collected in Los Angeles. Although nuScenes is not originally designed for traffic regulation analysis and lacks rule-level annotations, we manually reviewed front-camera images to identify samples where vehicle behavior is meaningfully influenced by traffic rules. Similarly, we sampled from our deployment logs in Los Angeles under comparable criteria.

\section{Experiments}

\subsection{Experimental Setup}
\noindent\textbf{Metrics.} 
We evaluate each model’s decision-making by assessing its ability to determine the compliance and safety of candidate actions in each scenario. For each action $a \in \mathcal{A}_{\text{cand}}$, the model outputs a predicted compliance indicator $l_c \in \{0,1\}$ and safety indicator $l_s \in \{0,1\}$. \textit{Compliance accuracy (Compl.)} and \textit{safety accuracy (Safety)} are computed as the percentage of actions for which the predicted compliance and safety indicators match the ground truth labels.

\noindent\textbf{Baseline Models.} 
We compare DriveReg with several state-of-the-art LLMs and VLMs~\cite{liu2024deepseekv3, guo2025deepseekr1, dubey2024llama, bai2023qwen, hurst2024gpt}. These baselines do not incorporate external traffic regulations but may reflect some rule understanding from pretraining. Their performance without explicit regulation input is denoted as \textit{Compl.\scriptsize{(-R)}} and \textit{Safety\scriptsize{(-R)}}. To assess the impact of our TRR module, we also evaluate these models with retrieved traffic rules of TRR, reported as \textit{Compl.\scriptsize{(+R)}} and \textit{Safety\scriptsize{(+R)}}.

\noindent \textbf{Implementation Details.}
We adopt ``text-embedding-ada-002'' as $\mathcal{E}$ and ``paraphrase-MiniLM-L6-v2'' as $\mathcal{E'}$. $\tau_p$ is set to 0.28 and $K=5$ for FAISS similarity search. Experiments are conducted over 5 random seeds. Additional settings are included in the extended version.

\begin{table*}[t]
    \centering
    \scriptsize
    \setlength{\tabcolsep}{4.5pt}
    \begin{tabular}{l|cccc|cccc}
    \toprule
    \multirow{2}{*}{Method}  & 
    \multicolumn{4}{c|}{Hypothesized Scenarios - Normal} & 
    \multicolumn{4}{c}{Hypothesized Scenarios - Hard} \\
    &  Compl.\scriptsize{(-R)} & Compl.\scriptsize{(+R)} & Safety\scriptsize{(-R)} & Safety\scriptsize{(+R)} & 
       Compl.\scriptsize{(-R)}& Compl.\scriptsize{(-R)} & Safety\scriptsize{(-R)} & Safety\scriptsize{(+R)} \\
    \midrule
    DeepSeek-V3 \cite{liu2024deepseekv3} & $0.87\pm 0.02$ & $0.92\pm 0.03$ & $0.90\pm 0.02$ & $0.92\pm 0.03$ & $0.82\pm 0.02$ & $0.92\pm 0.04$ & $0.84\pm 0.03$ & $0.84\pm 0.03$ \\
    DeepSeek-R1 \cite{guo2025deepseekr1}& $0.92\pm 0.02$ & $0.93\pm 0.02$ & $0.93\pm 0.04$ & $0.95\pm 0.01$ & $0.85\pm 0.01$ & $0.96\pm 0.03$ & $0.86\pm 0.05$ & $0.90\pm 0.03$ \\
    Llama3.3-70b \cite{dubey2024llama} & $0.90\pm 0.04$ & $0.91\pm 0.02$ & $0.93\pm 0.02$ & $0.94\pm 0.04$ & $0.81\pm 0.03$ & $0.93\pm 0.02$ & $0.86\pm 0.03$ & $0.90\pm 0.03$ \\
    Qwen2.5-72b \cite{bai2023qwen} & $0.93\pm 0.02$ & $0.94\pm 0.01$ & $0.92\pm 0.04$ & $0.95\pm 0.02$ & $0.88\pm 0.02$ & $0.95\pm 0.04$ & $0.85\pm 0.04$ & $0.86\pm 0.03$ \\
    GPT-4o \cite{hurst2024gpt} & $0.91\pm 0.02$ & $0.93\pm 0.03$ & $0.94\pm 0.01$ & $0.97\pm 0.02$ & $0.86\pm 0.03$ & $0.95\pm 0.01$ & $0.88\pm 0.02$ & $0.90\pm 0.02$ \\

    \textbf{DriveReg (Ours)} &  -- & $\textbf{0.98}\pm \textbf{0.01}$ & -- & $\textbf{0.98}\pm \textbf{0.01}$ &  -- & $\textbf{0.99}\pm \textbf{0.02}$ &  -- & $\textbf{0.94}\pm \textbf{0.02}$ \\

    \bottomrule
    \end{tabular}
    \caption{Average compliance and safety accuracy across three cities (Boston, Los Angeles, and Singapore) under DriveReg-Hypothesized driving scenarios. (-R) indicates baseline models relying solely on pretrained knowledge, while (+R) indicates that models provided with relevant traffic regulations retrieved by the TRR agent. }
    \label{tab:nuScenes_results}
\end{table*}

\subsection{Main Results}

\noindent \textbf{Hypothesized Scenarios Results.} 
We evaluated our framework’s performance on DriveReg hypothesized scenarios, with results presented in Table~\ref{tab:nuScenes_results}. Regulations retrieved via TRR improve performance across most models, particularly in hard cases, underscoring the effectiveness of our TRR agent. We observe that the improvement is notably evident in scenarios involving region-specific traffic laws, such as Boston’s requirement that drivers must remain in the right lane unless passing or preparing for a left turn, which are often overlooked by pretrained models. These results reveal the limitations of relying solely on general world knowledge encoded in foundation models, highlighting the need for explicit rule grounding in decision-making systems.

The DriveReg framework, incorporating the TRR Agent and rule-aware Reasoning Agent, achieves the highest overall performance. With access to regulatory input, DriveReg reaches 98\% compliance and 98\% safety accuracy in normal scenarios (F1: 97.6\%, p-value: 4.2e-03), and 99\% compliance and 94\% safety accuracy in hard scenarios (F1: 94.6\%, p-value: 3.5e-03). In particular, for hard cases, DriveReg improves compliance accuracy from 86\%, as obtained by the underlying GPT-4o model, to 99\% when enhanced with TRR and TRA. These results demonstrate that TRA, utilizing CoT reasoning, effectively interprets and applies retrieved traffic regulations to guide decision-making in scenarios requiring precise legal understanding.

\noindent \textbf{Real-World Scenarios Results.}
Table~\ref{tab:realworld_city} compares the performance of our DriveReg framework to the baseline GPT-4o on real-world scenarios across the three cities. Our model consistently outperforms the baseline in both compliance and safety metrics. Notably, GPT-4o achieves higher safety than compliance scores, suggesting VLMs capture general safety patterns (e.g., avoiding pedestrians or maintaining distance) better than precise legal constraints. In contrast, compliance requires accurate interpretation of traffic laws, which our framework addresses through retrieval-augmented reasoning. We also observe that some failure cases come from perception issues, such as misidentifying solid lane markings as broken, highlighting the importance of fine-grained visual understanding in urban environments.

Examples of output from our framework are shown in Fig.~\ref{fig:real_world_boston_singapore_la}. In (a), we present a scenario in Boston where the vehicle approaches a crosswalk. Here, the framework correctly identifies that accelerating forward is ``compliant but not safe'', which aligns with the common-sense guideline to use caution when approaching crosswalks, even when they are clear. In (b), we demonstrate the framework's ability to handle region-specific regulations: The ego vehicle attempts to turn right at a red light, an action that is illegal in Singapore but often permitted in the U.S. Our model correctly outputs ``non-compliant'', aligning with local traffic regulations. These examples highlight the framework's capacity to reason about both safety and region-specific laws. Additional results and video demonstrations are included in the extended version.

\begin{table}[t]
  \centering
  \scriptsize
  \begin{tabular}{l|cc|cc|cc}
    \toprule
    \multirow{2}{*}{Method} & \multicolumn{2}{c|}{Boston} & \multicolumn{2}{c|}{Los Angeles} & \multicolumn{2}{c}{Singapore} \\
    & Comp. & Safety & Comp. & Safety & Comp. & Safety \\
    \midrule
    GPT-4o & 0.86 & 0.91 & 0.91 & \textbf{0.93} & 0.84 & 0.89 \\
      \textbf{DriveReg (Ours)} & \textbf{0.94} & \textbf{0.92} & \textbf{0.95} & \textbf{0.93} & \textbf{0.91} & \textbf{0.93} \\
    \bottomrule
  \end{tabular}
  \caption{Comparison with GPT-4o on compliance and safety accuracy for real-world scenarios.}
  \label{tab:realworld_city}
\end{table}

\begin{table}[h]
\centering
\scriptsize
\begin{tabular}{c|cccc}
\toprule
Method & Best Match 25 & QLM & LLaDa TRE$^{\dagger}$ & \textbf{TRR (Ours)}\\
\midrule
Accuracy & 0.60 & 0.50 & 0.55 & \textbf{1.00} \\
\bottomrule
\end{tabular}
\caption{Traffic regulation retrieval accuracy comparison on 20 scenarios using top-5 retrieval. $^{\dagger}$: LLaDA TRE is reproduced following the original paper~\cite{li2024drivinglargelanguagemodel}.}
\label{tab:retrieval_comparison}
\end{table}

\begin{table}[t]
\centering
\scriptsize
\begin{tabular}{l|cc}
\toprule
Method & High-Level Comp. & High-Level Safety \\
\midrule
Agent-Driver~\cite{mao2024languageagentautonomousdriving} & 0.91 & 0.86 \\
\textbf{DriveReg (Ours)} & \textbf{0.98} &\textbf{ 0.97} \\
\bottomrule
\end{tabular}
\caption{High-level decision-making evaluation on the Boston real-world data split.}
\label{tab:agentdriver_comparison}
\end{table}

\noindent \textbf{Retrieval Method Comparison.}
We compare our TRR approach against traditional information retrieval methods on 20 Boston-Hard scenarios using top-5 retrieval: Best Match 25 (BM25) \cite{robertson2009probabilistic}, Query Likelihood Model (QLM) \cite{ponte2017language}, and the Traffic Rule Extractor (TRE) in LLaDa~\cite{li2024drivinglargelanguagemodel}. Accuracy is measured as the successful extraction of all ground truth regulations. As shown in Table~\ref{tab:retrieval_comparison}, our semantic similarity-based TRR Agent achieves superior retrieval accuracy, significantly outperforming keyword-based search methods.

\noindent \textbf{Comparison with Agent-Driver.} 
We further evaluate DriveReg against Agent-Driver~\cite{mao2024languageagentautonomousdriving}. Since Agent-Driver outputs trajectories without explicit compliance or safety indicators, we assess its reasoning performance by comparing its high-level plan outputs against a randomly selected action from DriveReg’s compliant-safe action set $\mathcal{A}^*$. As shown in Table~\ref{tab:agentdriver_comparison}, DriveReg outperforms Agent-Driver in both compliance and safety, indicating a stronger adherence to traffic rules.

\subsection{Real-World Testing}
\begin{figure}[t]
    \centering
    \includegraphics[width=\linewidth]{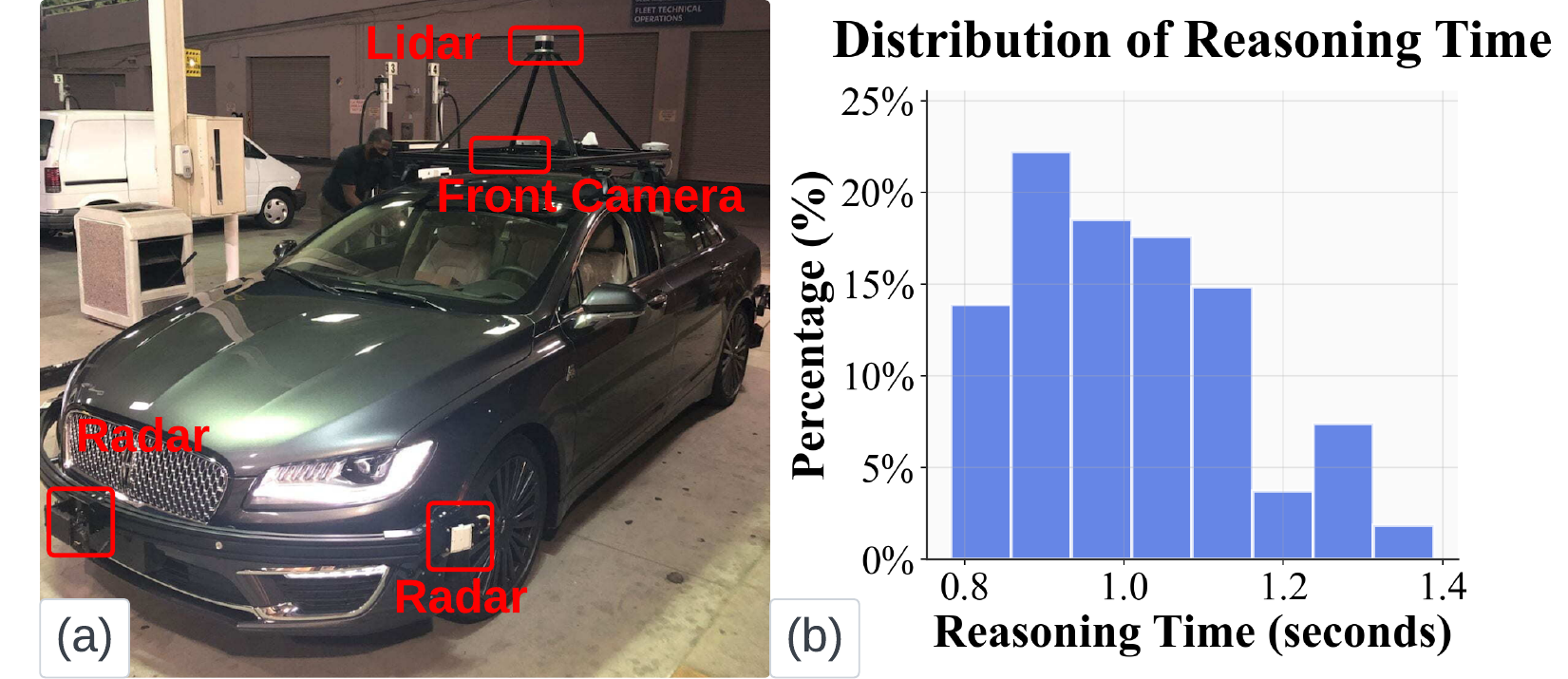}
    \caption{(a) The vehicle used for deployment. (b) Inference time distribution of the Reasoning Agent when additional reasoning steps are omitted for real-time efficiency.}
    \label{fig:realcar}
\end{figure}

To validate our framework in a real-world setting, we conducted deployment testing on an autonomous vehicle, as shown in Fig. \ref{fig:realcar} (a), in Los Angeles. For safety reasons, a human driver operated the vehicle while our system ran in parallel, fully functional but without actual control over the vehicle. This setup allowed us to evaluate the system's decision-making capabilities in real-time without compromising safety.
We tested the system in urban environments involving dynamic traffic signals, intersections, and interactions with vulnerable road users. The framework reliably retrieved relevant rules and produced safe, compliant decisions. Fig.~\ref{fig:real_world_boston_singapore_la} shows two examples: (c) a pedestrian crossing with a green light, and (d) a ``No Turn on Red'' intersection, both handled correctly by the system.

In addition to decision-making accuracy, we assess the inference time of DriveReg to evaluate real-time feasibility as shown in Fig.~\ref{fig:realcar} (b). Using GPT-4o, detailed reasoning outputs average 2 seconds per decision, while shorter outputs with only compliance/safety labels and rule numbers reduce latency to around 1 second, which is suitable for high-level decision-making in most driving scenarios.

\section{Conclusion}
We introduce an LLM-driven, traffic regulation-aware decision-making framework for autonomous driving that integrates a Traffic Rules Retrieval Agent and a Reasoning Agent. We also introduce a traffic rules scenario dataset to evaluate models on their ability to understand and reason over traffic regulations.  Experiments conducted on both hypothesized and real-world scenarios demonstrate the strong performance of our approach and its adaptability across different regions. The framework shows significant potential to enhance the interpretability and trustworthiness of autonomous driving systems. Future work will focus on extending the evaluation to a wider range of regions and refining the framework for broader real-world applicability.

\section*{Acknowledgment}
This work was supported by the U.S. Department of Transportation Federal Highway Administration (FHWA) Center of Excellence on New Mobility and Automated Vehicles.

\bibliography{aaai2026}

@misc{li2023knowledgedrivenautonomousdriving,
title={Towards Knowledge-driven Autonomous Driving},
author={Xin Li and Yeqi Bai and Pinlong Cai and Licheng Wen and Daocheng Fu and Bo Zhang and Xuemeng Yang and Xinyu Cai and Tao Ma and Jianfei Guo and Xing Gao and Min Dou and Yikang Li and Botian Shi and Yong Liu and Liang He and Yu Qiao},
year={2023},
eprint={2312.04316},
archivePrefix={arXiv},
primaryClass={[cs.RO](http://cs.ro/)},
url={https://arxiv.org/abs/2312.04316},
}

@article{Bae2020FiniteSM,
title={Finite State Machine based Vehicle System for Autonomous Driving in Urban Environments},
author={Sanghyeon Bae and Sunghyeon Joo and Jungwon Pyo and Jae-Seong Yoon and Kwanghee Lee and Tae-Yong Kuc},
journal={2020 20th International Conference on Control, Automation and Systems (ICCAS)},
year={2020},
pages={1181-1186},
url={https://api.semanticscholar.org/CorpusID:227278676}
}

@INPROCEEDINGS{9030183,
  author={Tadewos, Tadewos G. and Shamgah, Laya and Karimoddini, Ali},
  booktitle={2019 IEEE 58th Conference on Decision and Control (CDC)},
  title={Automatic Safe Behaviour Tree Synthesis for Autonomous Agents},
  year={2019},
  volume={},
  number={},
  pages={2776-2781},
  doi={10.1109/CDC40024.2019.9030183}}

@INPROCEEDINGS{9304549,
  author={Maierhofer, Sebastian and Rettinger, Anna-Katharina and Mayer, Eva Charlotte and Althoff, Matthias},
  booktitle={2020 IEEE Intelligent Vehicles Symposium (IV)},
  title={Formalization of Interstate Traffic Rules in Temporal Logic},
  year={2020},
  volume={},
  number={},
  pages={752-759},
  doi={10.1109/IV47402.2020.9304549}}

@misc{li2024drivinglargelanguagemodel,
      title={Driving Everywhere with Large Language Model Policy Adaptation}, 
      author={Boyi Li and Yue Wang and Jiageng Mao and Boris Ivanovic and Sushant Veer and Karen Leung and Marco Pavone},
      year={2024},
      eprint={2402.05932},
      archivePrefix={arXiv},
      primaryClass={cs.RO},
      url={https://arxiv.org/abs/2402.05932}, 
}

@misc{mao2024languageagentautonomousdriving,
      title={A Language Agent for Autonomous Driving}, 
      author={Jiageng Mao and Junjie Ye and Yuxi Qian and Marco Pavone and Yue Wang},
      year={2024},
      eprint={2311.10813},
      archivePrefix={arXiv},
      primaryClass={cs.CV},
      url={https://arxiv.org/abs/2311.10813}, 
}

@misc{lewis2021retrieval,
      title={Retrieval-Augmented Generation for Knowledge-Intensive NLP Tasks}, 
      author={Patrick Lewis and Ethan Perez and Aleksandra Piktus and Fabio Petroni and Vladimir Karpukhin and Naman Goyal and Heinrich Küttler and Mike Lewis and Wen-tau Yih and Tim Rocktäschel and Sebastian Riedel and Douwe Kiela},
      year={2021},
      eprint={2005.11401},
      archivePrefix={arXiv},
      primaryClass={cs.CL},
      url={https://arxiv.org/abs/2005.11401}, 
}

@InProceedings{pmlr-v162-borgeaud22a,
  title = 	 {Improving Language Models by Retrieving from Trillions of Tokens},
  author =       {Borgeaud, Sebastian and Mensch, Arthur and Hoffmann, Jordan and Cai, Trevor and Rutherford, Eliza and Millican, Katie and Van Den Driessche, George Bm and Lespiau, Jean-Baptiste and Damoc, Bogdan and Clark, Aidan and De Las Casas, Diego and Guy, Aurelia and Menick, Jacob and Ring, Roman and Hennigan, Tom and Huang, Saffron and Maggiore, Loren and Jones, Chris and Cassirer, Albin and Brock, Andy and Paganini, Michela and Irving, Geoffrey and Vinyals, Oriol and Osindero, Simon and Simonyan, Karen and Rae, Jack and Elsen, Erich and Sifre, Laurent},
  booktitle = 	 {Proceedings of the 39th International Conference on Machine Learning},
  pages = 	 {2206--2240},
  year = 	 {2022},
  editor = 	 {Chaudhuri, Kamalika and Jegelka, Stefanie and Song, Le and Szepesvari, Csaba and Niu, Gang and Sabato, Sivan},
  volume = 	 {162},
  series = 	 {Proceedings of Machine Learning Research},
  month = 	 {17--23 Jul},
  publisher =    {PMLR},
  url = 	 {https://proceedings.mlr.press/v162/borgeaud22a.html},
}

@misc{li2023pretraining,
      title={Pre-training on Synthetic Driving Data for Trajectory Prediction}, 
      author={Yiheng Li and Seth Z. Zhao and Chenfeng Xu and Chen Tang and Chenran Li and Mingyu Ding and Masayoshi Tomizuka and Wei Zhan},
      year={2023},
      eprint={2309.10121},
      archivePrefix={arXiv},
      primaryClass={cs.CV}
}

@InProceedings{wei2024editable,
      title={Editable Scene Simulation for Autonomous Driving via Collaborative LLM-Agents}, 
      author={Yuxi Wei and Zi Wang and Yifan Lu and Chenxin Xu and Changxing Liu and Hao Zhao and Siheng Chen and Yanfeng Wang},
      booktitle={Proceedings of the IEEE/CVF Conference on Computer Vision and Pattern Recognition (CVPR)},
      month={June},
      year={2024},
}

@inproceedings{malla2023drama,
  title={DRAMA: Joint Risk Localization and Captioning in Driving},
  author={Malla, Srikanth and Choi, Chiho and Dwivedi, Isht and Choi, Joon Hee and Li, Jiachen},
  booktitle={Proceedings of the IEEE/CVF Winter Conference on Applications of Computer Vision},
  pages={1043--1052},
  year={2023}
}

@article{sima2023drivelm,
  title={DriveLM: Driving with Graph Visual Question Answering},
  author={Sima, Chonghao and Renz, Katrin and Chitta, Kashyap and Chen, Li and Zhang, Hanxue and Xie, Chengen and Luo, Ping and Geiger, Andreas and Li, Hongyang},
  journal={arXiv preprint arXiv:2312.14150},
  year={2023}
}

@misc{nakano2022webgpt,
      title={WebGPT: Browser-assisted question-answering with human feedback}, 
      author={Reiichiro Nakano and Jacob Hilton and Suchir Balaji and Jeff Wu and Long Ouyang and Christina Kim and Christopher Hesse and Shantanu Jain and Vineet Kosaraju and William Saunders and Xu Jiang and Karl Cobbe and Tyna Eloundou and Gretchen Krueger and Kevin Button and Matthew Knight and Benjamin Chess and John Schulman},
      year={2022},
      eprint={2112.09332},
      archivePrefix={arXiv},
      primaryClass={cs.CL},
      url={https://arxiv.org/abs/2112.09332}, 
}

@article{bansal2018chauffeurnet,
  title={Chauffeurnet: Learning to drive by imitating the best and synthesizing the worst},
  author={Bansal, Mayank and Krizhevsky, Alex and Ogale, Abhijit},
  journal={arXiv preprint arXiv:1812.03079},
  year={2018}
}

@article{yuan2024evolutionary,
  title={Evolutionary decision-making and planning for autonomous driving based on safe and rational exploration and exploitation},
  author={Yuan, Kang and Huang, Yanjun and Yang, Shuo and Zhou, Zewei and Wang, Yulei and Cao, Dongpu and Chen, Hong},
  journal={Engineering},
  volume={33},
  pages={108--120},
  year={2024},
  publisher={Elsevier}
}

@article{mao2023gpt,
  title={Gpt-driver: Learning to drive with gpt},
  author={Mao, Jiageng and Qian, Yuxi and Zhao, Hang and Wang, Yue},
  journal={arXiv preprint arXiv:2310.01415},
  year={2023}
}

@inproceedings{tang2023personalized,
  title={Personalized Decision-Making and Control for Automated Vehicles Based on Generative Adversarial Imitation Learning},
  author={Tang, Xinyue and Yuan, Kang and Li, Shangwen and Yang, Shuo and Zhou, Zewei and Huang, Yanjun},
  booktitle={2023 IEEE 26th International Conference on Intelligent Transportation Systems (ITSC)},
  pages={4806--4812},
  year={2023},
  organization={IEEE}
}

@article{kiran2021deep,
  title={Deep reinforcement learning for autonomous driving: A survey},
  author={Kiran, B Ravi and Sobh, Ibrahim and Talpaert, Victor and Mannion, Patrick and Al Sallab, Ahmad A and Yogamani, Senthil and P{\'e}rez, Patrick},
  journal={IEEE Transactions on Intelligent Transportation Systems},
  volume={23},
  number={6},
  pages={4909--4926},
  year={2021},
  publisher={IEEE}
}

@article{wang2021decision,
  title={Decision making framework for autonomous vehicles driving behavior in complex scenarios via hierarchical state machine},
  author={Wang, Xuanyu and Qi, Xudong and Wang, Ping and Yang, Jingwen},
  journal={Autonomous Intelligent Systems},
  volume={1},
  pages={1--12},
  year={2021},
  publisher={Springer}
}

@article{li2021metadrive,
  title={Metadrive: Composing diverse driving scenarios for generalizable reinforcement learning},
  author={Li, Quanyi and Peng, Zhenghao and Feng, Lan and Zhang, Qihang and Xue, Zhenghai and Zhou, Bolei},
  journal={IEEE Transactions on Pattern Analysis and Machine Intelligence},
  year={2022}
}

@article{li2023scenarionet,
  title={ScenarioNet: Open-Source Platform for Large-Scale Traffic Scenario Simulation and Modeling},
  author={Li, Quanyi and Peng, Zhenghao and Feng, Lan and Liu, Zhizheng and Duan, Chenda and Mo, Wenjie and Zhou, Bolei},
  journal={Advances in Neural Information Processing Systems},
  year={2023}
}

@article{han2024foundation,
  title={Foundation intelligence for smart infrastructure services in transportation 5.0},
  author={Han, Xu and Meng, Zonglin and Xia, Xin and Liao, Xishun and He, Yueshuai and Zheng, Zhaoliang and Wang, Yutong and Xiang, Hao and Zhou, Zewei and Gao, Letian and others},
  journal={IEEE Transactions on Intelligent Vehicles},
  year={2024},
  publisher={IEEE}
}

@article{zhao2023autonomous,
  title={Autonomous driving system: A comprehensive survey},
  author={Zhao, Jingyuan and Zhao, Wenyi and Deng, Bo and Wang, Zhenghong and Zhang, Feng and Zheng, Wenxiang and Cao, Wanke and Nan, Jinrui and Lian, Yubo and Burke, Andrew F},
  journal={Expert Systems with Applications},
  pages={122836},
  year={2023},
  publisher={Elsevier}
}

@inproceedings{xiao2021rule,
  title={Rule-based optimal control for autonomous driving},
  author={Xiao, Wei and Mehdipour, Noushin and Collin, Anne and Bin-Nun, Amitai Y and Frazzoli, Emilio and Tebbens, Radboud Duintjer and Belta, Calin},
  booktitle={Proceedings of the ACM/IEEE 12th International Conference on Cyber-Physical Systems},
  pages={143--154},
  year={2021}
}

@inproceedings{manas2022robust,
  title={Robust Traffic Rules and Knowledge Representation for Conflict Resolution in Autonomous Driving.},
  author={Manas, Kumar and Zwicklbauer, Stefan and Paschke, Adrian},
  booktitle={RuleML+ RR (Companion)},
  year={2022}
}

@inproceedings{sun2022lawbreaker,
  title={LawBreaker: An approach for specifying traffic laws and fuzzing autonomous vehicles},
  author={Sun, Yang and Poskitt, Christopher M and Sun, Jun and Chen, Yuqi and Yang, Zijiang},
  booktitle={Proceedings of the 37th IEEE/ACM International Conference on Automated Software Engineering},
  pages={1--12},
  year={2022}
}

@misc{wen2024enhancingsociallyawarerobotnavigation,
      title={Enhancing Socially-Aware Robot Navigation through Bidirectional Natural Language Conversation}, 
      author={Congcong Wen and Yifan Liu and Geeta Chandra Raju Bethala and Zheng Peng and Hui Lin and Yu-Shen Liu and Yi Fang},
      year={2024},
      eprint={2409.04965},
      archivePrefix={arXiv},
      primaryClass={cs.RO},
      url={https://arxiv.org/abs/2409.04965}, 
}

@article{zhang2024large,
  title={Large motion model for unified multi-modal motion generation},
  author={Zhang, Mingyuan and Jin, Daisheng and Gu, Chenyang and Hong, Fangzhou and Cai, Zhongang and Huang, Jingfang and Zhang, Chongzhi and Guo, Xinying and Yang, Lei and He, Ying and others},
  journal={arXiv preprint arXiv:2404.01284},
  year={2024}
}

@inproceedings{caesar2020nuscenes,
  title={nuscenes: A multimodal dataset for autonomous driving},
  author={Caesar, Holger and Bankiti, Varun and Lang, Alex H and Vora, Sourabh and Liong, Venice Erin and Xu, Qiang and Krishnan, Anush and Pan, Yu and Baldan, Giancarlo and Beijbom, Oscar},
  booktitle={Proceedings of the IEEE/CVF conference on computer vision and pattern recognition},
  pages={11621--11631},
  year={2020}
}

@article{johnson2017billion,
  title={Billion-scale similarity search with GPUs},
  author={Johnson, Jeff and Douze, Matthijs and J{\'e}gou, Herv{\'e}},
  journal={arXiv preprint arXiv:1702.08734},
  year={2017}
}

@inproceedings{Li2024VioHawk,
  author    = {Zhongrui Li and Jiarun Dai and Zongan Huang and Nianhao You and Yuan Zhang and Min Yang},
  title     = {{VioHawk}: Detecting Traffic Violations of Autonomous Driving Systems through Criticality-Guided Simulation Testing},
  booktitle = {Proceedings of the ACM SIGSOFT International Symposium on Software Testing and Analysis (ISSTA)},
  year      = {2024}
}

@article{Deng2025TARGET,
  author  = {Yao Deng and Zhi Tu and Jiaohong Yao and Mengshi Zhang and others},
  title   = {{TARGET}: LLM-Guided Generation of Traffic-Rule Test Scenarios for Autonomous Vehicles},
  journal = {IEEE Transactions on Software Engineering},
  year    = {2025},
  note    = {Early Access}
}

@article{liu2024deepseekv3,
  title={Deepseek-v3 technical report},
  author={Liu, Aixin and Feng, Bei and Xue, Bing and Wang, Bingxuan and Wu, Bochao and Lu, Chengda and Zhao, Chenggang and Deng, Chengqi and Zhang, Chenyu and Ruan, Chong and others},
  journal={arXiv preprint arXiv:2412.19437},
  year={2024}
}

@article{guo2025deepseekr1,
  title={Deepseek-r1: Incentivizing reasoning capability in llms via reinforcement learning},
  author={Guo, Daya and Yang, Dejian and Zhang, Haowei and Song, Junxiao and Zhang, Ruoyu and Xu, Runxin and Zhu, Qihao and Ma, Shirong and Wang, Peiyi and Bi, Xiao and others},
  journal={arXiv preprint arXiv:2501.12948},
  year={2025}
}

@article{dubey2024llama,
  title={The llama 3 herd of models},
  author={Dubey, Abhimanyu and Jauhri, Abhinav and Pandey, Abhinav and Kadian, Abhishek and Al-Dahle, Ahmad and Letman, Aiesha and Mathur, Akhil and Schelten, Alan and Yang, Amy and Fan, Angela and others},
  journal={arXiv e-prints},
  pages={arXiv--2407},
  year={2024}
}

@article{bai2023qwen,
  title={Qwen technical report},
  author={Bai, Jinze and Bai, Shuai and Chu, Yunfei and Cui, Zeyu and Dang, Kai and Deng, Xiaodong and Fan, Yang and Ge, Wenbin and Han, Yu and Huang, Fei and others},
  journal={arXiv preprint arXiv:2309.16609},
  year={2023}
}

@article{hurst2024gpt,
  title={Gpt-4o system card},
  author={Hurst, Aaron and Lerer, Adam and Goucher, Adam P and Perelman, Adam and Ramesh, Aditya and Clark, Aidan and Ostrow, AJ and Welihinda, Akila and Hayes, Alan and Radford, Alec and others},
  journal={arXiv preprint arXiv:2410.21276},
  year={2024}
}

@article{robertson2009probabilistic,
  title={The probabilistic relevance framework: BM25 and beyond},
  author={Robertson, Stephen and Zaragoza, Hugo and others},
  journal={Foundations and Trends{\textregistered} in Information Retrieval},
  volume={3},
  number={4},
  pages={333--389},
  year={2009},
  publisher={Now Publishers, Inc.}
}

@inproceedings{ponte2017language,
  title={A language modeling approach to information retrieval},
  author={Ponte, Jay M and Croft, W Bruce},
  booktitle={ACM SIGIR Forum},
  volume={51},
  number={2},
  pages={202--208},
  year={2017},
  organization={ACM New York, NY, USA}
}

@article{wilson2023argoverse,
  title={Argoverse 2: Next generation datasets for self-driving perception and forecasting},
  author={Wilson, Benjamin and Qi, William and Agarwal, Tanmay and Lambert, John and Singh, Jagjeet and Khandelwal, Siddhesh and Pan, Bowen and Kumar, Ratnesh and Hartnett, Andrew and Pontes, Jhony Kaesemodel and others},
  journal={arXiv preprint arXiv:2301.00493},
  year={2023}
}

@inproceedings{nuplan,
  title={Towards learning-based planning: The nuPlan benchmark for real-world autonomous driving},
  author={Karnchanachari, Napat and Geromichalos, Dimitris and Tan, Kok Seang and Li, Nanxiang and Eriksen, Christopher and Yaghoubi, Shakiba and Mehdipour, Noushin and Bernasconi, Gianmarco and Fong, Whye Kit and Guo, Yiluan and others},
  booktitle={2024 IEEE International Conference on Robotics and Automation (ICRA)},
  pages={629--636},
  year={2024},
  organization={IEEE}
}

@article{MEHDIPOUR2023110692,
title = {Formal methods to comply with rules of the road in autonomous driving: State of the art and grand challenges},
journal = {Automatica},
volume = {152},
pages = {110692},
year = {2023},
issn = {0005-1098},
doi = {https://doi.org/10.1016/j.automatica.2022.110692},
url = {https://www.sciencedirect.com/science/article/pii/S0005109822005568},
author = {Noushin Mehdipour and Matthias Althoff and Radboud Duintjer Tebbens and Calin Belta}
}

@article{kubica2022AVLL,
    author = {Kubica, María Lubomira},
    title = {Autonomous Vehicles and Liability Law},
    journal = {The American Journal of Comparative Law},
    volume = {70},
    pages = {i39-i69},
    year = {2022},
    month = {08},
    issn = {0002-919X},
    doi = {10.1093/ajcl/avac015}
}

@book{USDOT2016FAVP,
  title={Federal automated vehicles policy: Accelerating the next revolution in roadway safety},
  author={National Highway Traffic Safety Administration and others},
  year={2016},
  publisher={US Department of Transportation}
}

@inproceedings{huang2023gameformer,
  title={Gameformer: Game-theoretic modeling and learning of transformer-based interactive prediction and planning for autonomous driving},
  author={Huang, Zhiyu and Liu, Haochen and Lv, Chen},
  booktitle={Proceedings of the IEEE/CVF International Conference on Computer Vision},
  pages={3903--3913},
  year={2023}
}

@inproceedings{huang2025gen,
  title={Gen-drive: Enhancing diffusion generative driving policies with reward modeling and reinforcement learning fine-tuning},
  author={Huang, Zhiyu and Weng, Xinshuo and Igl, Maximilian and Chen, Yuxiao and Cao, Yulong and Ivanovic, Boris and Pavone, Marco and Lv, Chen},
  booktitle={2025 IEEE International Conference on Robotics and Automation (ICRA)},
  pages={3445--3451},
  year={2025},
  organization={IEEE}
}


\newpage
\makeatletter
\@ifundefined{isChecklistMainFile}{
  \newif\ifreproStandalone
  \reproStandalonetrue
}{
  \newif\ifreproStandalone
  \reproStandalonefalse
}
\makeatother
\clearpage

\ifreproStandalone
\documentclass[letterpaper]{article}
\usepackage{aaai2026}  
\setlength{\pdfpagewidth}{8.5in}
\setlength{\pdfpageheight}{11in}
\usepackage{times}
\usepackage{helvet}
\usepackage{courier}
\usepackage{xcolor}
\usepackage{booktabs}
\usepackage{colortbl}
\usepackage{multirow}
\usepackage{graphicx}
\usepackage{cite}
\usepackage{amsmath}
\usepackage{natbib}  
\frenchspacing

\begin{document}
\fi
\setlength{\leftmargini}{20pt}
\makeatletter\def\@listi{\leftmargin\leftmargini \topsep .5em \parsep .5em \itemsep .5em}
\def\@listii{\leftmargin\leftmarginii \labelwidth\leftmarginii \advance\labelwidth-\labelsep \topsep .4em \parsep .4em \itemsep .4em}
\def\@listiii{\leftmargin\leftmarginiii \labelwidth\leftmarginiii \advance\labelwidth-\labelsep \topsep .4em \parsep .4em \itemsep .4em}\makeatother

\setcounter{secnumdepth}{2}
\renewcommand\thesubsection{\thesection.\arabic{subsection}}
\renewcommand\labelenumi{\thesubsection.\arabic{enumi}}

\newcounter{checksubsection}
\newcounter{checkitem}[checksubsection]

\newcommand{\checksubsection}[1]{%
  \refstepcounter{checksubsection}%
  \paragraph{\arabic{checksubsection}. #1}%
  \setcounter{checkitem}{0}%
}

\newcommand{\checkitem}{%
  \refstepcounter{checkitem}%
  \item[\arabic{checksubsection}.\arabic{checkitem}.]%
}
\newcommand{\question}[2]{\normalcolor\checkitem #1 #2 \color{blue}}
\newcommand{\ifyespoints}[1]{\makebox[0pt][l]{\hspace{-15pt}\normalcolor #1}}

\renewcommand{\thesection}{\Alph{section}}
\renewcommand{\thetable}{S\arabic{table}}
\renewcommand{\thefigure}{S\arabic{figure}}
\renewcommand{\theequation}{S\arabic{equation}}

\setcounter{figure}{0}
\setcounter{table}{0}
\setcounter{equation}{0}
\setcounter{section}{0}

\twocolumn[
\begin{center}
\Large \textbf{Driving with Regulation: Trustworthy and Interpretable Decision-Making for \\
\Large Autonomous Driving with Retrieval-Augmented Reasoning}\\
\Large Supplementary Material
\end{center}
\vspace{0.5cm}
]

\section{DriveReg Scenario Dataset Details}
\label{sec:regulation_sources}

This section presents detailed descriptions of the regulatory sources used in constructing the \textbf{DriveReg Scenario Dataset} across Boston (MA), Los Angeles (CA), and Singapore. Our dataset includes diverse types of legal documents and traffic-related materials to support comprehensive and region-specific decision-making evaluation.

\subsection{Types of Traffic Regulation Sources}

The regulatory documents used in DriveReg are drawn from five source types.

\textit{City-Level Traffic Regulation}: Set by local governments, rules to address specific needs such as parking, speed limits, and lane usage to manage local traffic and ensure safety.

\textit{State-Level Traffic Law}: Laws regulate vehicle operations and ensure road safety, established by state legislatures and enforced statewide.

\textit{State-Level Driving Manual}: Published by each state's DMV, this manual details state traffic laws and safe driving practices. It includes driving safety guidelines in the form of text and illustrated images.

\textit{State-Level Court Cases}: Judicial rulings on traffic-related cases clarify laws and influence enforcement. 

\textit{Traffic Norms}: Widely recognized behaviors that drivers follow to ensure smooth and safe road interactions. These norms are essential for autonomous vehicles to align with human driving behaviors and societal expectations. This paper does not focus on building a repository of records for such norms, but we will use examples to illustrate that our framework still applies.

\subsection{City-Specific Regulation Sources}

\noindent\textbf{Boston (Massachusetts, USA)}

\textit{City-Level Regulations}: Boston City Traffic Rules and Regulations - Article IV (One-Way Regulations) and Article V (Operation of Vehicles)

\textit{State-Level Laws}: Massachusetts General Laws - Chapter 89 (Law of the Road), Sections 1-12
    
\textit{State-Level Manual}: Massachusetts Driver's Manual (Handbook) - Chapter 4 (Rules of the Road) and Chapter 5 (Special Driving Situations)
    
\textit{Court Cases}: Twelve selected Massachusetts traffic violation cases
    
\textit{Traffic Norms}: Ten selected US driving norms without regulation violations

\

\noindent\textbf{Los Angeles (California, USA)}

\textit{City-Level Regulations}: Los Angeles City Traffic Rules and Regulations  - Chapter VIII (Traffic), Divisions A through W
    
\textit{State-Level Laws}: California Vehicle Code - Division 11 (Rules of the Road)
    
\textit{State-Level Manual}: California Driver's Manual (Handbook) - Section 6 (Navigating the Roads), Section 7 (Laws and Rules of the Road), Section 8 (Safe Driving), Section 12 (Driver Safety)
    
\textit{Court Cases}: Ten selected California traffic violation cases
    
\textit{Traffic Norms}: Ten selected US driving norms without regulation violations

\

\noindent \textbf{Singapore (City-State in Southeast Asia)}

\textit{State-Level Laws}: Singapore Road Traffic Act 1961
    
\textit{State-Level Manual}: Singapore Basic Theory of Driving - Part B (Signs and Signals, Traffic Rules and Regulations, Code of Conduct on the Road, Parts and Controls of a Car, Work Zone, Driving in Special Zones, Driving in Tunnels, Autonomous Vehicle) and Part C (Useful Information)
    
\textit{Court Cases}: Twelve selected Singapore traffic violation cases
    
\textit{Traffic Norms}: Ten selected Singapore driving norms without regulation violations

\subsection{Traffic Document Preprocessing}

To ensure consistent and structured access to regulatory information, we implemented a comprehensive preprocessing pipeline that accommodates the diverse formats and sources of traffic-related materials while preserving the original legal language and intended meaning of the documents.

\textbf{Web-Based Document Extraction}: For documents such as state laws and court cases that are accessible only via web interfaces without direct download options, we employed automated web crawlers to extract their content. This ensures access to the most up-to-date versions while preserving their original structural organization and formatting.

\textbf{PDF Handbook Processing}: For driver's manuals and handbooks typically distributed in PDF format, we implemented a specialized processing approach that separates textual content from visual elements. The text portions are extracted and processed, while images are systematically labeled and their descriptive labels are strategically reinserted into the text flow. This ensures that visual information critical to understanding traffic rules is preserved in textual form.

\textbf{Format Standardization}: All extracted content is converted into a standardized Markdown format. This format balances human readability with machine-parsable structure, enabling efficient indexing and retrieval during rule search and reasoning.

\textbf{Quality Assurance}: After automated processing, each document is manually reviewed to correct formatting inconsistencies, misaligned headings, indentation errors, and artifacts from OCR or parsing. This final step ensures high fidelity between the processed documents and their original regulatory intent.

The resulting corpus provides a well-structured representation of regulatory texts, suitable for research on traffic rule understanding, retrieval, and decision-making.

\begin{figure}[h]
    \centering
    \includegraphics[width=0.98\linewidth]{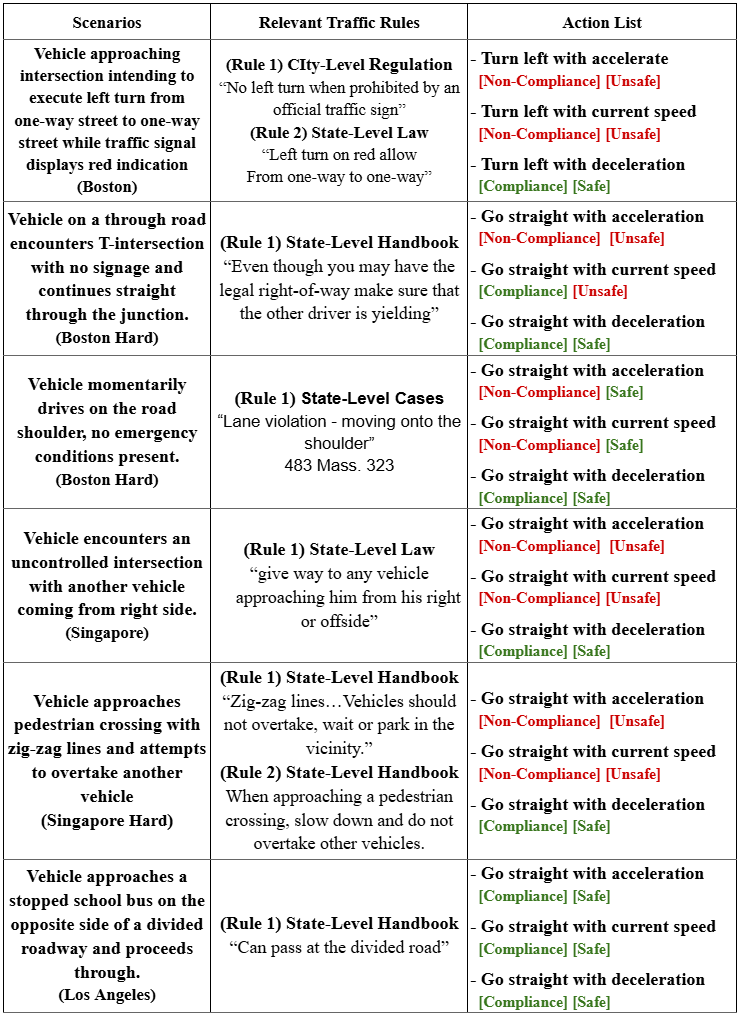}
    \caption{Additional hypothesized scenario examples from the DriveReg Scenarios Dataset, showing scenario descriptions, relevant traffic rules, and action-level compliance and safety annotations.}
    \label{fig:additional_hypo_examples}
\end{figure}

\subsection{Hypothesized Scenarios Details}

To ensure a comprehensive evaluation of traffic rule understanding, we designed hypothesized scenarios to systematically cover all major regulatory topics found in the documents of each region. The category distribution for Boston is provided in the main paper (Table 1). In Table~\ref{tab:hypo_scenarios_classification_la} and Table~\ref{tab:hypo_scenarios_classification_singapore}, we present the corresponding distributions for Los Angeles and Singapore. As shown, our hypothesized scenarios offer broad and balanced coverage across key traffic rule categories, supporting meaningful comparison and generalization across jurisdictions.

\begin{table}[h]
\centering
\footnotesize
\setlength{\tabcolsep}{1mm}
\begin{tabular}{lclc}
\toprule
\textbf{Category} & \textbf{Count} & \textbf{Category} & \textbf{Count} \\
\midrule
Right-of-Way Rules     & 20 & Special Driving Situations & 13 \\
Intersections          & 15 & Rules for Passing         & 4  \\
Pedestrians            & 15 & Traffic Signs \& Signals  & 11  \\
Pavement Markings      & 10 & Roadway Construction      & 4  \\
Service Vehicles       & 8  &  Hard Cases         & 20 \\
\bottomrule
\end{tabular}
\caption{Distribution of hypothesized scenarios in Los Angeles.}
\label{tab:hypo_scenarios_classification_la}
\end{table}

\begin{table}[h]
\centering
\footnotesize
\setlength{\tabcolsep}{1mm}
\begin{tabular}{lclc}
\toprule
\textbf{Category} & \textbf{Count} & \textbf{Category} & \textbf{Count} \\
\midrule
Right-of-Way Rules     & 22 & Special Driving Situations & 11 \\
Intersections          & 17 & Rules for Passing         & 3  \\
Pedestrians            & 15 & Traffic Signs \& Signals  & 8  \\
Pavement Markings      & 13 & Roadway Construction      & 3  \\
Service Vehicles       & 8  &  Hard Cases         & 20 \\
\bottomrule
\end{tabular}
\caption{Distribution of hypothesized scenarios in Singapore.}
\label{tab:hypo_scenarios_classification_singapore}
\end{table}

In Fig. \ref{fig:additional_hypo_examples}, we present additional examples of hypothesized scenarios from the DriveReg Scenarios Dataset. Each row includes (1) a natural language description of the traffic situation, (2) the relevant traffic rules retrieved from city-level, state-level, or court case sources, and (3) a list of candidate actions with compliance and safety annotations. These examples illustrate the diverse regulatory contexts represented in our dataset, covering region-specific laws, ambiguous intersections, overtaking restrictions, and school bus regulations. By including multi-source rule references and fine-grained action-level labels, these examples further demonstrate the dataset’s coverage, interpretability, and utility for evaluating rule-aware decision-making.

\subsection{Additional Real-world Scenarios Details}

We collect real-world driving scenarios from two sources: the nuScenes dataset \cite{caesar2020nuscenes} and our deployment logs in Los Angeles. The nuScenes dataset is a large-scale, real-world autonomous driving dataset containing annotated sensor data from urban driving scenes in Boston and Singapore. Although it was not originally designed for traffic regulation reasoning, we manually reviewed front-camera images to identify samples that are meaningfully influenced by traffic rules, such as intersections, stop signs, pedestrian crossings, and yield situations. From these selected scenes, we annotated compliance and safety labels for all candidate actions in the predefined action set, and additionally identified the relevant traffic regulations that apply to each scenario. For Los Angeles, we collected real-world data from our in-vehicle deployment during on-road testing. The vehicle was driven in a variety of urban environments, including major roads, secondary streets, intersections (with and without traffic signals), and pedestrian-dense areas. We used camera-recorded footage to select frames where traffic regulation awareness is critical. Similar to the nuScenes-based samples, we annotated each scenario with compliance and safety labels, as well as the relevant traffic rules.

To maintain a balanced dataset and avoid redundancy, we omitted some overly repetitive scenarios, particularly those that share identical regulatory contexts. In total, we curated 140 real-world scenarios, with the distribution by city summarized in Table~\ref{tab:realworld_sample_counts}. These real-world scenarios cover a wide range of traffic rule types, including \textit{Right-of-Way Rules}, \textit{Intersections}, \textit{Pedestrians}, \textit{Pavement Markings}, and \textit{Traffic Signs \& Signals}. In contrast, categories such as \textit{Service Vehicles} and \textit{Special Driving Situations} are less represented, as these cases are rarely encountered in urban driving data. Unlike the hypothesized scenarios, which were constructed to ensure comprehensive rule coverage, the real-world scenarios are limited to what naturally appears in recorded driving scenes. As a result, the observed rules tend to reflect more common and frequently encountered regulatory situations.

\begin{table}[h]
\centering
\footnotesize
\begin{tabular}{l c}
\toprule
\textbf{City} & \textbf{Number of Samples} \\
\midrule
Boston     & 57 \\
Singapore  & 39 \\
Los Angeles  & 44 \\
\midrule
\textbf{Total} & \textbf{140} \\
\bottomrule
\end{tabular}
\caption{Number of real-world scenarios in the DriveReg Scenario Dataset from nuScenes and Los Angeles.}
\label{tab:realworld_sample_counts}
\end{table}

\section{Implementation Details}

\subsection{Environment Analysis}

\begin{figure}[h]
    \centering
    \includegraphics[width=\linewidth]{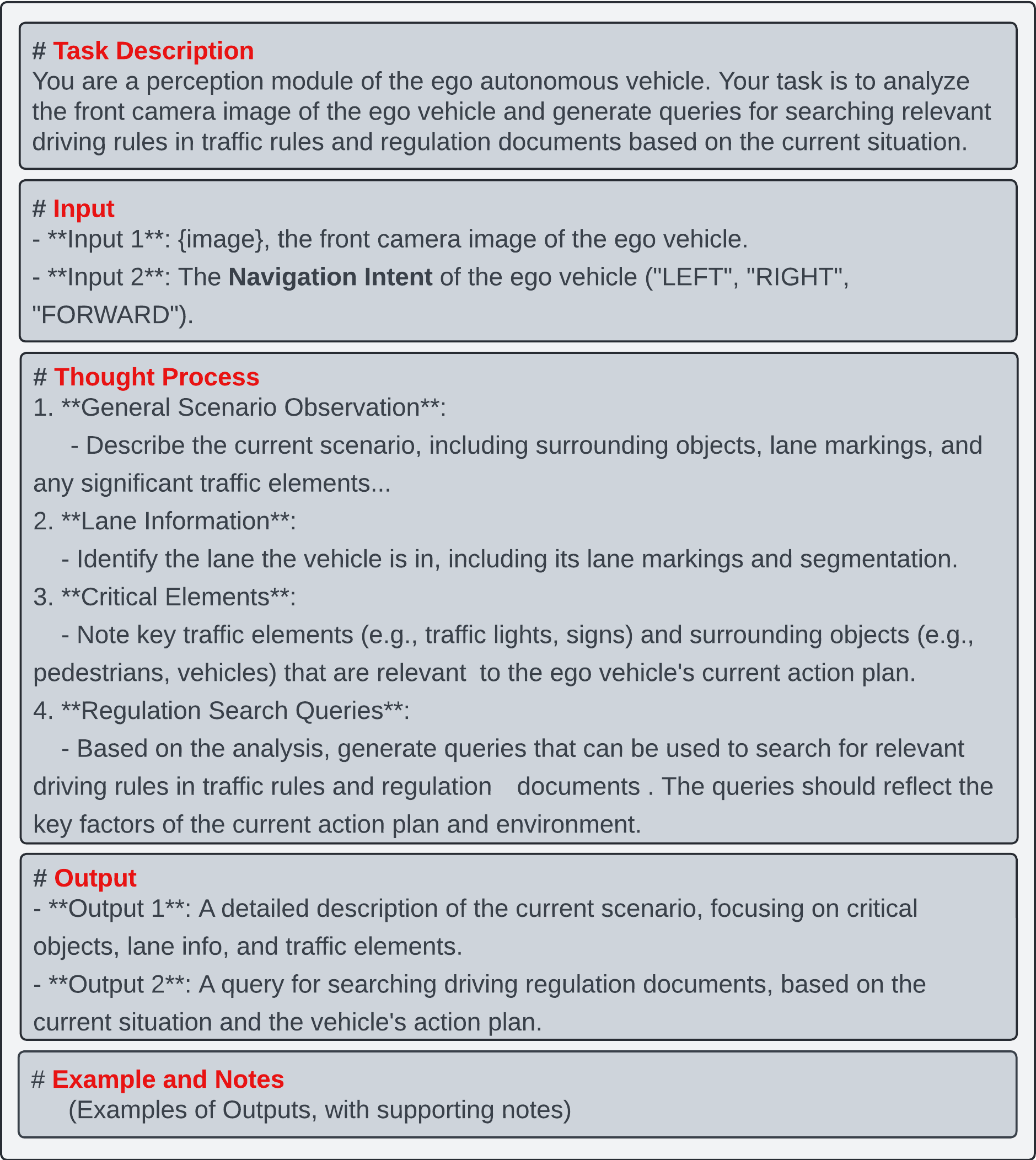}
    \caption{System prompt of Environment Analysis}
    \label{fig:systemprompt_1}
\end{figure}

We use a Vision-Language Model (VLM), guided by a structured system prompt (Fig.~\ref{fig:systemprompt_1}), as the perception module for environment analysis. The module receives a front-facing camera image and the ego vehicle's navigation intent as input, and outputs a structured scene description along with a natural language query for retrieving relevant traffic regulations.

This module is implemented using GPT-4o with few-shot Chain-of-Thought prompting. While our current design leverages a vision-language foundation model, the environment analysis component is modular and can be replaced with any structured perception pipeline capable of producing scene-level summaries and regulation-aware query cues.

\subsection{FAISS-Based Traffic Rule Retrieval}
Our Traffic Regulation Retrieval (TRR) Agent employs Facebook AI Similarity Search (FAISS) \cite{johnson2017billion} for efficient dense vector retrieval.

FAISS is a library for efficient similarity search and clustering of dense vectors, specifically designed for high-dimensional embeddings. It provides optimized algorithms for approximate nearest neighbor (ANN) search that can handle millions of vectors with sub-linear query complexity. FAISS offers scalability through both CPU and GPU implementations with efficient memory management for large-scale vector databases. The library supports multiple indexing methods, including flat search (exact), IVF (Inverted File), HNSW (Hierarchical Navigable Small World), and PQ (Product Quantization), with configurable parameters allowing optimization for either retrieval speed or precision based on application requirements.

For our traffic regulation retrieval task, we utilize FAISS with cosine similarity as the distance metric, which is well-suited for normalized text embeddings. The system performs approximate nearest neighbor search over pre-computed embeddings of regulatory documents, enabling real-time retrieval of relevant traffic rules during autonomous driving scenario analysis.

\subsection{Baseline Model Details}

\textbf{Large Language and Vision-Language Models:} We evaluate DriveReg against several state-of-the-art models spanning different architectures and parameter scales:

DeepSeek-V3 \cite{liu2024deepseekv3}: A 671B parameter Mixture-of-Experts (MoE) model with 37B activated parameters, representing one of the largest open-source language models. It employs sparse expert routing for efficient scaling and demonstrates strong performance in mathematical reasoning, code generation, and multi-step logical reasoning tasks.
    
\textit{DeepSeek-R1} \cite{guo2025deepseekr1}: A 671B parameter reasoning-focused variant optimized through advanced chain-of-thought training and RLHF techniques. The model excels in multi-hop reasoning, mathematical problem-solving, and systematic problem decomposition across STEM domains.
    
\textit{Llama3.3-70B} \cite{dubey2024llama}: Meta's 70B parameter decoder-only transformer featuring robust instruction-following capabilities and comprehensive safety alignment through constitutional AI principles. The model demonstrates strong performance in conversational AI, content generation, and analytical reasoning.

\textit{Qwen2.5-72B} \cite{bai2023qwen}: Alibaba's 72B parameter model with enhanced multilingual capabilities (30+ languages) and advanced reasoning performance. It incorporates mixture-of-experts knowledge distillation and excels in code generation, technical analysis, and multi-turn conversations.
    
\textit{GPT-4o} \cite{hurst2024gpt}: OpenAI's multimodal foundation model with integrated vision and language understanding through unified transformer architecture. It processes both textual and visual inputs for comprehensive scene analysis, visual reasoning, and cross-modal understanding tasks.

These baseline models are evaluated under two conditions: without external regulation access to assess their inherent traffic rule understanding from pretraining, and with retrieved traffic regulations to evaluate the effectiveness of our TRR module.

    


\

\noindent \textbf{Baseline Retrieval Methods:} To validate the effectiveness of our embedding-based retrieval pipeline, we compare against traditional information retrieval baselines:

\textit{BM25} \cite{robertson2009probabilistic}: Best Matching 25 is a probabilistic ranking function that combines term frequency (TF) and inverse document frequency (IDF) with document length normalization. For a query $Q$ containing terms $q_1, q_2, \ldots, q_n$ and document $D$, the BM25 score is computed as:
\begin{align}
\text{BM25}(Q, D) &= \sum_{i=1}^{n} \text{IDF}(q_i) \cdot \\
&\quad \frac{f(q_i, D) \cdot (k_1 + 1)}{f(q_i, D) + k_1 \cdot \left(1 - b + b \cdot \frac{|D|}{\text{avgdl}}\right)} \nonumber,
\end{align}
where $f(q_i, D)$ is the frequency of term $q_i$ in document $D$, $|D|$ is the document length, and $\text{avgdl}$ is the average document length in the collection. The parameters $k_1$ (1.2) and $b$ (0.75) control term frequency saturation and document length normalization, respectively. The IDF component is defined as:
\begin{equation}
\text{IDF}(q_i) = \log \frac{N - n(q_i) + 0.5}{n(q_i) + 0.5},
\end{equation}
where $N$ is the total number of documents and $n(q_i)$ is the number of documents containing term $q_i$.
    
\textit{Query Likelihood Model (QLM)} \cite{ponte2017language}: QLM is a language modeling approach that estimates the probability of generating the query from each document's language model. The model ranks documents by computing $P(Q|D)$, the probability that document $D$ would generate query $Q$:
\begin{equation}
P(Q|D) = \prod_{i=1}^{n} P(q_i|D).
\end{equation}

To address the problem of zero probabilities for unseen terms, QLM typically employs Jelinek-Mercer smoothing:
\begin{equation}
P(q_i|D) = \lambda \cdot P_{ML}(q_i|D) + (1-\lambda) \cdot P_{ML}(q_i|C),
\end{equation}
where $P_{ML}(q_i|D) = \frac{f(q_i, D)}{|D|}$ is document-level probability, $P_{ML}(q_i|C) = \frac{\sum_{D' \in C} f(q_i, D')}{\sum_{D' \in C} |D'|}$ is collection-level probability, and $\lambda=0.4$ controls interpolation.

\subsection{Additional Experimental Setting}
We adopt GPT-4o in DriveReg with temperature = 0.1, top-p = 0.95, and a maximum token length of 2048. The same configurations are used for all baseline models to ensure a fair comparison. These hyperparameters, along with others detailed in the main paper, are selected empirically based on performance across different settings.






\section{Additional Results}
\subsection{Additional Hypothesized Scenarios Results}

\begin{table*}
\centering
\footnotesize
\renewcommand{\arraystretch}{1.15}
\setlength{\tabcolsep}{3pt}
\begin{tabular}{l|l|cccc|cccc}
\toprule[1.1pt]
\multirow{2}{*}{City} & \multirow{2}{*}{Method} & 
\multicolumn{4}{c|}{Hypothesized Scenarios - Normal} & 
\multicolumn{4}{c}{Hypothesized Scenarios - Hard} \\
& & Compl.(-R) & Compl.(+R) & Safety(-R) & Safety(+R) & 
    Compl.(-R) & Compl.(+R) & Safety(-R) & Safety(+R) \\
\midrule
\multirow{6}{*}{Boston}
&DeepSeek-V3 & 0.86 & 0.89 & 0.85 & 0.88 & 0.83 & 0.93 & 0.80 & 0.80 \\
&DeepSeek-R1 & 0.93 & 0.93 & 0.92 & 0.92 & 0.88 & 0.95 & 0.88 & 0.90 \\
&Llama3.3-70b & 0.88 & 0.89 & 0.91 & 0.93 & 0.80 & 0.93 & 0.88 & 0.88 \\
&Qwen2.5-72b & 0.90 & 0.91 & 0.88 & 0.89 & 0.88 & 0.95 & 0.83 & 0.85 \\
&GPT-4o & 0.90 & 0.93 & 0.88 & 0.93 & 0.85 & 0.95 & 0.88 & 0.88 \\
&\textbf{DriveReg (Ours)} & - &\textbf{0.99} & - & \textbf{0.99} & -&\textbf{1.00} & - &\textbf{0.90} \\
\midrule
\multirow{6}{*}{Los Angeles}
&DeepSeek-V3 & 0.87 & 0.92 & 0.92 & 0.95 & 0.83 & 0.90 & 0.88 & 0.88 \\
&DeepSeek-R1 & 0.91 & 0.91 & 0.95 & 0.95 & 0.85 & 0.95 & 0.88 & 0.88 \\
&Llama3.3-70b & 0.92 & 0.94 & 0.91 & 0.97 & 0.80 & 0.93 & 0.85 & 0.88 \\
&Qwen2.5-72b & 0.93 & 0.94 & 0.94 & 0.97 & 0.88 & 0.95 & 0.83 & 0.83 \\
&GPT-4o & 0.92 & 0.93 & 0.96 & 0.99 & 0.85 & 0.98 & 0.88 & 0.90 \\
&\textbf{DriveReg (Ours) }& - & \textbf{0.95} & - & \textbf{0.99} &  - & \textbf{0.95} &  - & \textbf{0.93} \\
\midrule
\multirow{6}{*}{Singapore}
&DeepSeek-V3 & 0.91 & 0.96 & 0.93 & 0.93 & 0.80 & 0.93 & 0.83 & 0.85 \\
&DeepSeek-R1 & 0.92 & 0.94 & 0.97 & 0.99 & 0.85 & 0.98 & 0.88 & 0.90 \\
&Llama3.3-70b & 0.91 & 0.92 & 0.98 & 0.98 & 0.83 & 0.93 & 0.90 & 0.93 \\
&Qwen2.5-72b & 0.93 & 0.97 & 0.94 & 0.98 & 0.88 & 0.95 & 0.88 & 0.90 \\
&GPT-4o & 0.91 & 0.94 & 0.97 & 0.98 & 0.88 & 0.95 & 0.88 & 0.93 \\
&\textbf{ DriveReg (Ours)} & - & \textbf{0.97} & - &  \textbf{1.00} & - & \textbf{0.98} & - & \textbf{0.93}\\
\bottomrule[1.1pt]
\end{tabular}
\caption{Compliance and safety accuracy on hypothesized driving scenarios across three cities in the DriveReg Scenarios Dataset. (-R) indicates baseline models relying solely on pretrained knowledge, while (+R) indicates that models are provided with relevant traffic regulations retrieved by the TRR agent.}
\label{tab:3city_results}
\end{table*}

Table~\ref{tab:3city_results} shows the detailed results for the individual cities: Boston, Los Angeles, and Singapore. These results correspond to the averages reported in Table~2 of the main paper. DriveReg consistently achieves the highest performance across all three cities and both scenario types. All methods show decreased performance on Hard scenarios compared to Normal scenarios, confirming that our hard case subset introduces greater complexity. This validates that the \textbf{Hard Scenarios are indeed challenging}, as intended by our scenario design. However, DriveReg maintains the smallest performance gap, indicating superior handling of complex regulatory situations. Moreover, the consistent improvement from the \textit{without retrieved regulation} (-R) to the \textit{with retrieved regulation} (+R) setting across all baselines and scenario types confirms the critical importance of accessing relevant regulatory information for accurate compliance and safety evaluation.

\begin{figure}[ht]
    \centering
    \medskip
    \includegraphics[width=\linewidth]{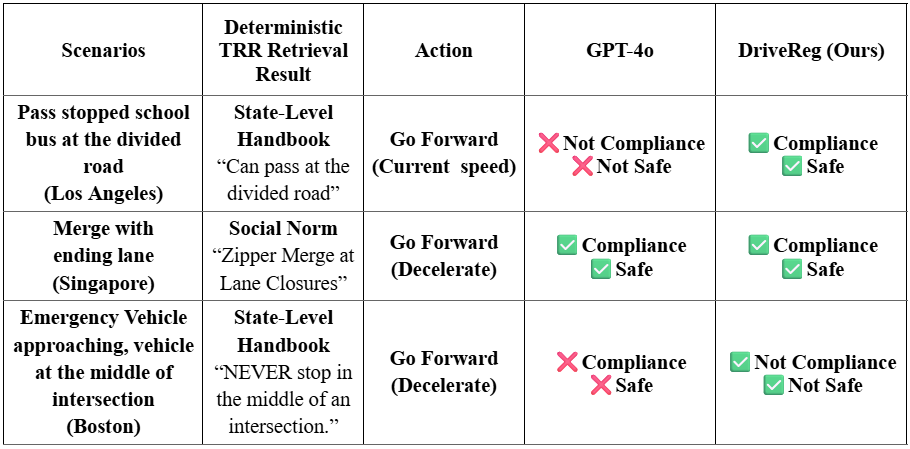}
    \caption{Comparison between GPT-4o and DriveReg on hypothesized scenarios from Los Angeles, Singapore, and Boston. DriveReg produces more accurate compliance and safety assessments by using retrieved traffic regulations.}
    \label{fig:qualitative_table_sup}
\end{figure}

We provide additional examples comparing DriveReg and GPT-4o in Fig.~\ref{fig:qualitative_table_sup}. While GPT-4o sometimes misjudges compliance or safety, DriveReg can provide correct assessments in those cases by grounding its decisions in retrieved traffic rules and structured reasoning.

\subsection{Additional Real-World Scenarios Results}

\begin{figure*}
    \centering
    \medskip
    \includegraphics[width=\linewidth]{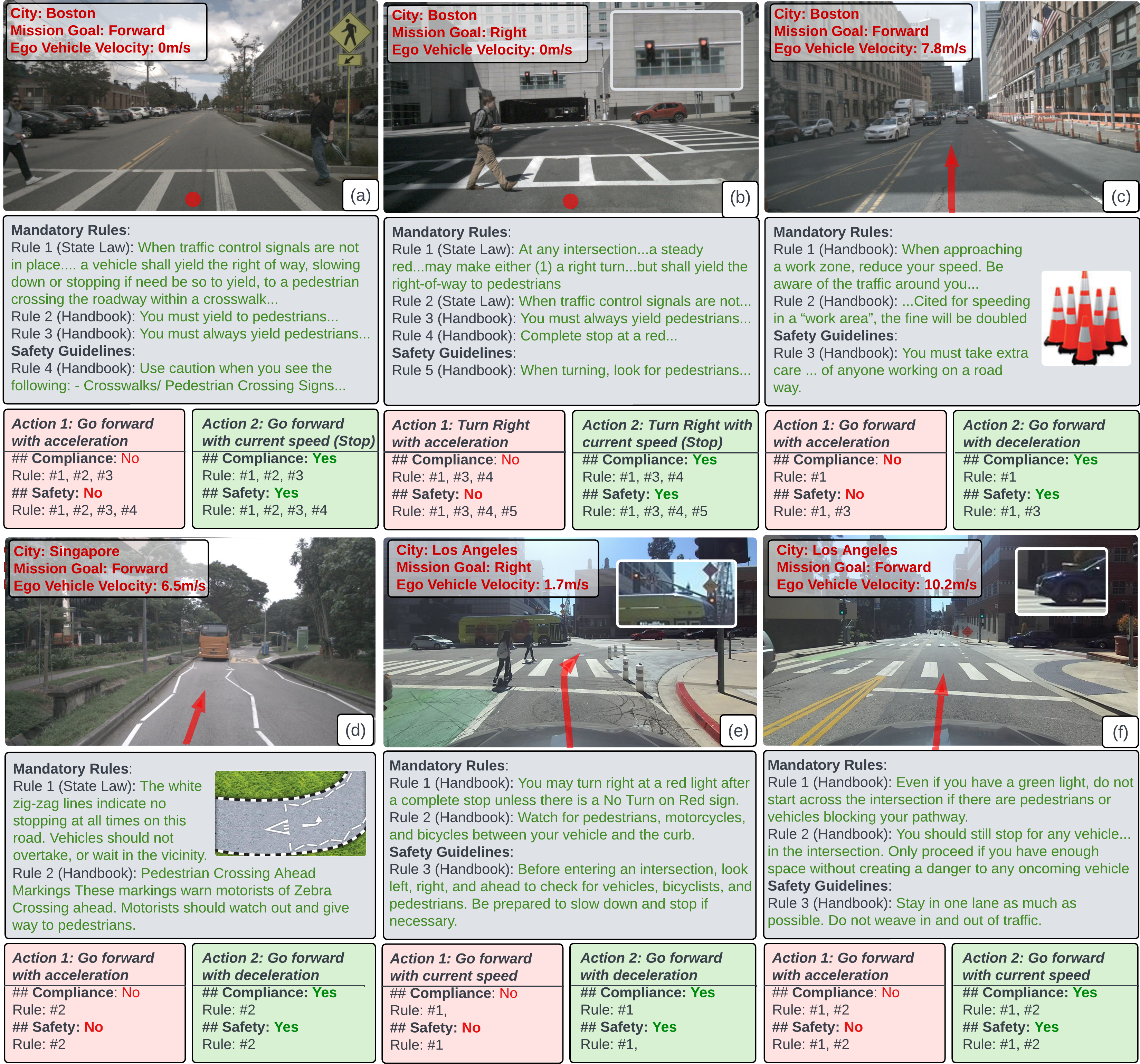}
    \caption{Inference results from the DriveReg (Real-World) Scenarios Dataset.}
    \label{fig:realworld_supp}
\end{figure*}

Fig.~\ref{fig:realworld_supp} presents additional examples of our framework’s output on real-world scenarios. In (a) and (b), we present two scenarios: a crosswalk without traffic control signals and a crosswalk with traffic control signals. Our framework correctly identifies this and retrieves the relevant traffic rules. These two examples demonstrate our model's ability to reason based on multiple environmental factors, accurately adjusting its rule-retrieval and evaluation according to changes in the scenario. 

In (b), we further demonstrate DriveReg's ability in cases where multiple traffic elements and rules must be considered simultaneously. In this scenario, the vehicle is turning right at a red traffic light where there is no “No Turn on Red” sign, making the turn legally permissible. However, a pedestrian is crossing the crosswalk in front of the vehicle, requiring the vehicle to yield. Thus, turning right without yielding is non-compliant with traffic regulations. As shown in the final output, our model successfully identifies this and outputs “non-compliant” for the case. 

In (c), we present a case where the ego vehicle is approaching a work zone and should reduce its speed, which our model successfully identifies, outputting the action “go forward with deceleration.” This is a scenario that previous rule-based methods struggle to handle, as they typically select only key rules due to the need for manually hand-crafting the rules, often omitting specific cases such as regulations for work zones. 

In (d), we present a scenario from Singapore featuring white zig-zag lines ahead of a pedestrian crossing. These markings are unique to regions like Singapore and indicate that vehicles must not stop or overtake in the area. Our framework successfully identifies these regulatory markings and retrieves the corresponding local traffic rules, demonstrating its ability to adapt to region-specific traffic signals. 

In (e) and (f), we show complex real-world cases in Los Angeles where multiple traffic rules must be considered simultaneously. In both cases, DriveReg successfully reasons over the relevant rules and produces the correct output.


\subsection{Human Evaluation of Explanation Quality}
To further assess the interpretability of DriveReg’s decisions, we conducted a human study evaluating the quality of its generated explanations. We invited 15 participants, each of whom reviewed 40 explanation-decision pairs sampled across all three cities (Boston, Los Angeles, Singapore) and both scenario types (Normal and Hard). For each pair, participants were shown the scenario description, the candidate action set, DriveReg’s compliance/safety judgment, and the corresponding Chain-of-Thought explanation. Participants then rated the explanation on a simple 1-5 quality scale, focusing on its consistency with the final decision and how helpful it was for understanding the compliance or safety judgment. DriveReg achieved an average score of \textbf{4.93}, demonstrating that its reasoning is clear, reliable, and well aligned with its decisions.

\end{document}
\fi

\end{document}